\documentclass{sigkddExp}
\usepackage{url}
\usepackage{graphics}
\usepackage{xspace}

\newcommand{\eg}{\textit{e.g.},\xspace} 
\newcommand{\ie}{\textit{i.e.},\xspace}
\newcommand{\etal}{et~al.}
\newcommand{\iv}{\textit{in vivo}}

\newcommand{\burst}{$\boxtimes$}

\begin{document}

\title{``In vivo'' spam filtering:\\A challenge problem for data mining}

\numberofauthors{1} \author{
  \alignauthor Tom Fawcett \\
  \affaddr{Hewlett-Packard Laboratories}\\
  \affaddr{1501 Page Mill Road}\\
  \affaddr{Palo Alto, CA  USA}\\
  \email{tom.fawcett@hp.com} }

\maketitle

\begin{abstract}
  Spam, also known as Unsolicited Commercial Email (UCE), is the bane
  of email communication.  Many data mining researchers have addressed
  the problem of detecting spam, generally by treating it as a static
  text classification problem.  True \textit{in vivo} spam filtering
  has characteristics that make it a rich and challenging domain for
  data mining.  Indeed, real-world datasets with these characteristics
  are typically difficult to acquire and to share.  This paper
  demonstrates some of these characteristics and argues that
  researchers should pursue \textit{in vivo} spam filtering as an
  accessible domain for investigating them.

\end{abstract}

\terms{spam, text classification, challenge problems, class skew, imbalanced
  data, cost-sensitive learning, data streams, concept drift}

\section{Introduction}

Spam, also known as Unsolicited Commercial Email (UCE) and Unsolicited Bulk
Email (UBE), is commonplace everywhere in email communication\footnote{The
  term ``spam'' is sometimes used loosely to mean any message broadcast to
  multiple senders (regardless of intent) or any message that is undesired.
  Here we intend the narrower, stricter definition: unsolicited commercial
  email sent to an account by a person unacquainted with the recipient.}.
Spam is a costly problem and many experts agree it is only getting worse
\cite{CranorLaMacchia:1998,Olsen:2002,Schneider:2003,Weinstein:2003,Gleick:2003}.
Because of the economics of spam and the difficulties inherent in stopping it,
it is unlikely to go away soon.

Many data mining and machine learning researchers have worked on spam
detection and filtering, commonly treating it as a basic text classification
problem.  The problem is popular enough that it has been the subject of a Data
Mining Cup contest \cite{DMC:2003} as well as numerous class projects.
Bayesian analysis has been very popular
\cite{Sahami98,Schneider03,Graham:2003,AndroutsopoulosEtal:2000a}, but
researchers have also used SVMs \cite{Kolcz01}, decisions trees
\cite{Carreras01}, memory and case-based reasoning
\cite{Sakkis:2001,Cunningham_wkshp:2003}, rule learning \cite{Provost:1999}
and even genetic programming \cite{Katirai:99}.

But researchers who treat spam filtering as an isolated text classification
task have only addressed a portion of the problem.  This paper argues that
real-world \iv\ spam filtering is a rich and challenging problem for data
mining.  By ``\textit{in vivo}'' we mean the problem as it is truly faced in
an operating environment, that is, by an on-line filter on a mail account
that receives realistic feeds of email over time, and serves a human user.  In
this context, spam filtering faces issues of skewed and changing class
distributions; unequal and uncertain error costs; complex text patterns; a
complex, disjunctive and drifting target concept; and challenges of
intelligent, adaptive adversaries.  Many real-world domains share these
characteristics and would benefit indirectly by work on spam filtering.

Improving spam filtering is a worthy goal in itself, but this paper
takes the (admittedly selfish) position that data mining researchers
should study the problem for the benefit of data mining.  It is
unclear whether spam filtering efforts could genuinely benefit from
data mining research.  On the other hand, one of the persistent
difficulties of research in many real-world domains is that of
acquiring and sharing datasets.  Most companies, for example, do not
release customer transaction data; we are aware of no public domain
datasets containing genuine fraudulent transactions for studying fraud
detection.  Even sharing such data between partner companies usually
requires formal non-disclosure agreements.  In other domains datasets
may still have copyright or privacy issues.  Few datasets involving
concept drift or changing class distributions are publicly available.
Without such datasets, the ability to replicate results and to compare
algorithm performance is hindered and progress on these research
topics will be impaired.  Spam data are easily accessible and
shareable, which makes spam filtering a good domain testbed for
investigating many of the same issues.

The remainder of the paper enumerates these research issues and describes how
they are manifested in \iv\ spam filtering.  The final section of the paper
discusses how researchers could begin exploring the domain.

\begin{figure*}[tb]
  \centering
  \begin{tabular}{cc}
    \epsfig{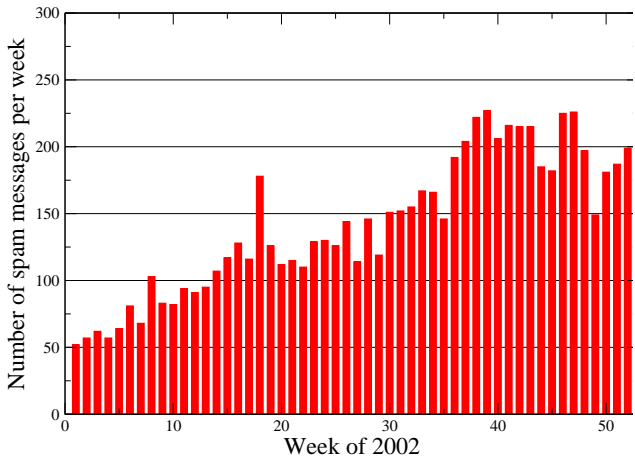} &
    \epsfig{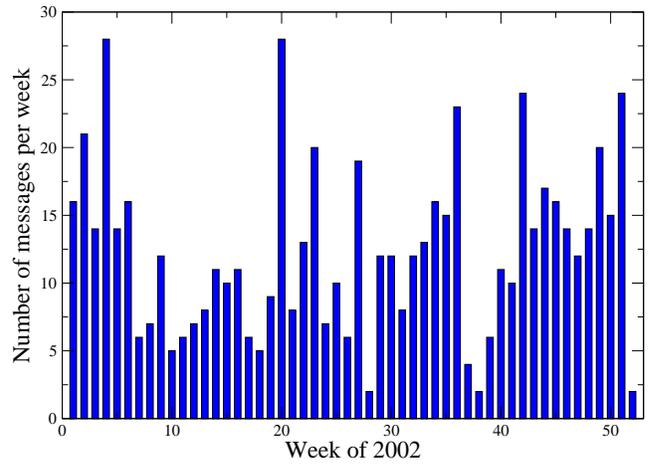}
    \\
    (a) Spam messages & (b) Legitimate messages \\
  \end{tabular}
  \caption{Weekly variation in message traffic, spam versus legitimate email}
  \vspace*{3ex}
  \label{fig:weekly_volume}
\end{figure*}

\begin{figure}[thb]
  \centering \epsfig{file=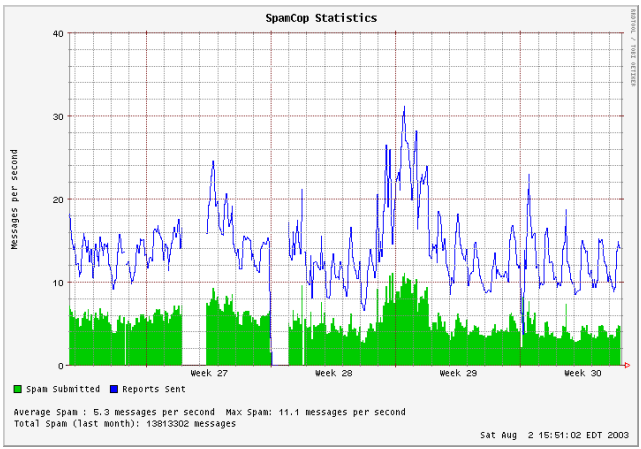,width=\columnwidth}
  \caption{SpamCop: Spam forwarded and reports sent}
  \label{fig:spamcop_month}
\end{figure}

\begin{figure}[thb]
  \centering \epsfig{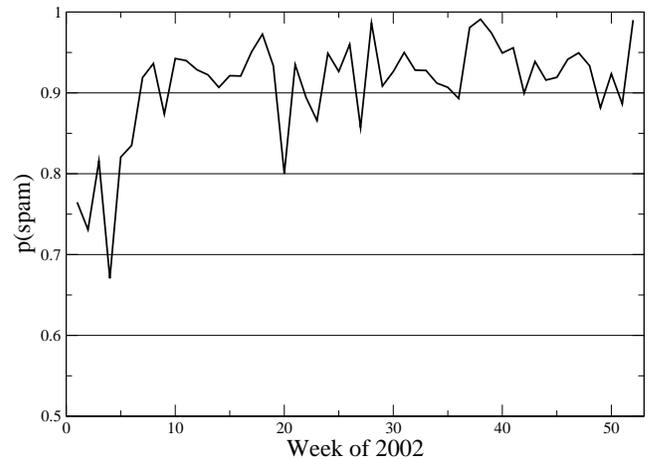}
  \caption{Drifting priors: weekly estimates of $p(\text{spam})$ taken
    from data in figure 1.}
  \label{fig:priors}
\end{figure}

\section{Challenges}

\subsection{Skewed and drifting class distributions}
\label{sect:class-skew}

Like most text classification domains, spam presents the problem of a skewed
class distribution, \ie\ the proportion of spam to legitimate email is uneven.
There are no generally agreed upon class priors for this problem.  G\'omez
Hidalgo \cite{Gomez02a} points out that the proportion of spam messages
reported in research datasets varies considerably, from 16.6\% to 88.2\%.
This may be simply because the proportion varies considerably from one
individual to another.  The amount of spam received depends on the email
address, the degree of exposure, the amount of time the address has been
public and the upstream filtering.  The amount of legitimate email received
similarly varies greatly from one individual to another.


Perhaps more importantly, spam varies over \emph{time} as well.  This
was demonstrated dramatically in 2002 when a large number of open
relays and open proxies were brought on-line in Asian countries,
primarily Korea and China.  Such a large new pool of unprotected
machines provided great opportunities for spammers, and soon email
servers throughout the world experienced a huge surge in the amount of
spam they forwarded and received.  The problem became so bad that for
a brief time all email from certain Asian countries was blocked
completely by some ISPs \cite{Delio:2002}.

In spite of claims that spam is generally increasing
\cite{CranorLaMacchia:1998,Olsen:2002,CAUBE.AU:2002}, the volume varies
considerably and non-monotonically on a daily or weekly scale.  Calculating
spam proportion even approximately is difficult.  Although some public spam
datasets are available (see Appendix A), we are aware of no personal email
datasets arranged over time, so it is difficult to match the two to establish
priors.  Nevertheless, using several datasets we can make a case that spam
priors change significantly over time.

\begin{figure*}[htb]
  \centering
  \begin{tabular}{cc}
    \epsfig{file=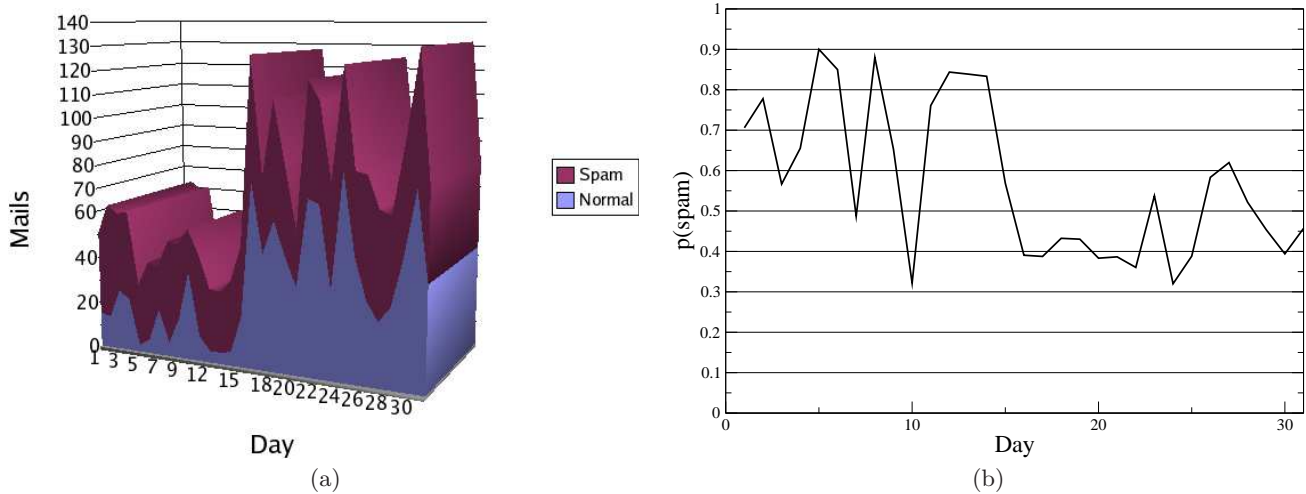,width=\columnwidth} &
    \epsfig{file=keide.eps,width=\columnwidth} \\
    (a) & (b) 
  \end{tabular}
  \caption{Email volume from Eide's trial,  (a) absolute volume, (b) resulting
    prior $p(\text{spam})$.}
  \label{fig:keide}
\end{figure*}

Figure~\ref{fig:weekly_volume}a shows a graph of spam volume received in 2002
by Paul Wouters of
Xtended~Internet\footnote{\url{http://spamarchive.xtdnet.nl/}}.  In 2002 the
spam volume was $146\pm 55$ messages per week, indicating a great deal of
variation in spite of its upward trend.  For most people, the volume of the
legitimate email received varies as well.  Figure~\ref{fig:weekly_volume}b
shows a graph of the number of legitimate messages saved by the author over
the weeks in 2002.  The volume is $12.3 \pm 6.4$ messages per week.


Figure~\ref{fig:spamcop_month} shows the volume of reports issued from
SpamCop's
website\footnote{\url{http://www.spamcop.net/spamstats.shtml}}
This graph also demonstrates some of spam's episodic nature.  SpamCop
is a service used by many people to filter spam and to submit reports
(complaints) to the originators of spam.  Both the amount of spam
submitted and the number of reports sent show clear episodic behavior.

These graphs show time variation in both the volume of spam and the volume of
legitimate email received, something that researchers have not generally
acknowledged.  Since the two sources of email---senders of spam and senders of
legitimate email---are independent parties with little in common, we can
expect their variation to be statistically uncorrelated, and the class priors
will vary over time.  No fixed prior will be correct.

How much could we expect class priors to vary?  If we assume that a user
received the spam shown in figure~\ref{fig:weekly_volume}a and the legitimate
email shown in figure~\ref{fig:weekly_volume}b, we can estimate the class
prior $p(\text{spam})$ simply as the proportion of weekly messages that are
spam.  Figure~\ref{fig:priors} shows a graph of this value, which ranges
between about .67 to .99.

A further demonstration of changing priors appears in Kristian Eide's study of
bayesian spam filters \cite{Eide:2003}.  In evaluating these filters he
measured the volume of spam and legitimate email he received over the course
of one month.  These volumes are graphed in figure~\ref{fig:keide}a, and the
computed daily spam prior is graphed in figure~\ref{fig:keide}b.  The prior
ranges from .32 to .9, showing greater variation than in
figure~\ref{fig:priors}, though the skew is not as high.

Variation in class priors may be problematic for researchers because it makes
solution superiority more difficult to establish.  A classifier that performs
better than another on a dataset with 80\% spam may perform worse on one with
40\% spam \cite{ProvostFawcettKohavi:98}.

Should researchers be concerned about these varying class priors?
This question is difficult to answer conclusively because it depends
on classifier performance as well as error cost assumptions (discussed
in section~\ref{sec:vary-uneq-error}).  But by employing the cost
curve framework of Drummond and Holte \cite{DrummondHolte:2000}, we
can answer a related question, \emph{How much of cost space is
  influenced by this variation?}  This question can be answered by
calculating
the span of the Probability Cost Function (PCF), which is the $x$ axis
of a cost curve.  The PCF ranges from zero to one and is a function of
the class prevalence and the ratio of misclassification error costs.
In the case of spam:
\begin{align*}
  \text{PCF}_{\text{spam}} & = \frac{p(\text{spam}) \cdot \text{cost}(FN)}{p(\text{spam})
    \cdot \text{cost}(FN) + p(\text{legit})\cdot \text{cost}(FP)}\\
  \intertext{If we assume that the cost of a false positive (that is, of
  classifying a legitimate message as spam) is about ten times that of
  a false negative, this reduces to}
  \text{PCF}_{\text{spam}} & = \frac{p(\text{spam})}{p(\text{spam}) + p(\text{legit}) \times
    10}\\
\end{align*}
Using the $p(\text{spam})$ range from figure~\ref{fig:keide}b, the PCF range
of interest for spam filtering is $.04 \leq \text{PCF}_{\text{spam}} \leq .47$.  Since the
entire PCF range is [0,1], this means nearly half of cost space is influenced
by this variation in priors.  Any classifier whose performance lies within
this 44\% could be a competitive solution.  This is a wide range, and it is
reasonable to expect classifier superiority to vary within it.

The purpose of this analysis is not to call into question the validity of
prior work, but to point out that changing class distributions are a reality
in this domain and their influence on solutions should be tested.  Conversely,
researchers investigating skewed and varying class distributions would do well
to study the spam filtering problem.

Exactly how a researcher should best track and adjust class priors is
an open question and will require research.  Time series work in
statistics should provide some strategies, for example, using an
exponentially decayed average of recent priors.  However researchers
estimate priors, they should acknowledge that priors vary and static
values are unrealistic.

\subsection{Unequal and uncertain error costs}
\label{sec:vary-uneq-error}

A further complication of \iv\ filtering is the asymmetry of error
costs.  Viewing the filter as a spam classifier, a spam message is a
positive instance and a legitimate message is a negative instance.
Judging a legitimate email to be spam (a false positive error) is
usually far worse than judging a spam email to be legitimate (a false
negative error).  A false negative simply causes slight irritation,
\ie\ the user sees an undesirable message.  A false positive can be
critical.  If spam is deleted permanently from a mail server, a false
positive can be very expensive since it means a (possibly important)
message has been discarded without a trace.  If spam is moved to a
low-priority mail folder for later human scanning, or if the address
is only used to receive low priority email, false positives may be
much more tolerable.

In an essay on developing a bayesian spam filter, Paul Graham
\cite{Graham:2003} describes the different errors in an insightful
comment:
\begin{quotation}
  \noindent False positives seem to me a different kind of error from false
  negatives.  Filtering rate is a measure of performance.  False positives I
  consider more like bugs.  I approach improving the filtering rate as
  optimization, and decreasing false positives as debugging.
\end{quotation}

Ken Schneider, CTO of the mail filtering company BrightMail, makes the
same point more starkly \cite{Schneider:2003}.  He argues that
filtering even a small amount of legitimate email defeats the purpose
of filtering because it forces the user to start reviewing the spam
folder for missed messages.  Even a single missed important message
may cause a user to reconsider the value of spam filtering.  This
argues for assigning a very high cost to false positive errors.

Regardless of the exact values, these asymmetric error costs must be
acknowledged and taken into account by any acceptable filtering
solution.  Judging a spam filtering system by accuracy (or,
equivalently, error rate) is unrealistic and misleading
\cite{ProvostFawcettKohavi:98}.  Some researchers have measured
precision and recall without questioning whether metrics for
information retrieval are appropriate for a filtering task.

Fortunately, most researchers have acknowledged these asymmetric costs, but
methods for dealing with them have been ad hoc.  The 2003 Data Mining Cup
Competition \cite{DMC:2003} required that learned classifiers have no more
than a 1\% false positive rate, but the organizers gave no justification for
this cut-off.  Graham \cite{Graham:2003} simply double-counted the tokens of
his legitimate email, essentially considering the cost of a false positive to
be twice that of a false negative.  Sahami \etal\ \cite{Sahami98} used a very
high probability threshold of $.999$ for classifying a message as spam.
Androutsopoulos \etal\ \cite{AndroutsopoulosEtal:2000b} performed more careful
experiments across cost ratios of 1, 10 and 100, exploring two orders of
magnitude of cost ratios.

These approaches suggest a deeper issue: true costs of filtering errors may
simply be unknown to the data mining researcher, or may be known only
approximately.  Only the end user will know the consequences of filtering
mistakes and be able to estimate error tradeoffs.  \textit{In vivo} filtering
requires flexibility of solutions: the user should be able to specify the
approximate costs (or relative severity) of the errors and the run-time filter
should accommodate.  Admittedly this requirement complicates research
evaluation since the superiority of an approach may not extend throughout a
cost range, and multiple experiments may have to be performed.

Such uncertainty is actually common in real-world domains, where
experts may have difficulty stating the exact cost of an erroneous
action, or the cost of the action may vary depending on external
circumstances.  This situation motivated development of a framework
based on ROC analysis for evaluating and managing classifiers when
error costs are uncertain \cite{ProvostFawcett:2001}.  In the case of
spam filtering, the uncertainty of error costs may not change
temporally but they do vary between users.  G\'omez Hidalgo
\cite{Gomez02a} used this framework for developing and evaluating spam
filtering solutions, and found it useful.  Drummond and Holte
\cite{DrummondHolte:2000} have also developed a cost curve framework
that extends ROC analysis and serves much the same purpose.  Whatever
technique is used for evaluating classifier performance, researchers
should be prepared to demonstrate a solution's performance over a
range of costs.

\subsection{Disjunctive and changing target concept}
\label{sect:concept_drift}

{ \setlength{\tabcolsep}{0mm}
  \begin{figure*}[tb]
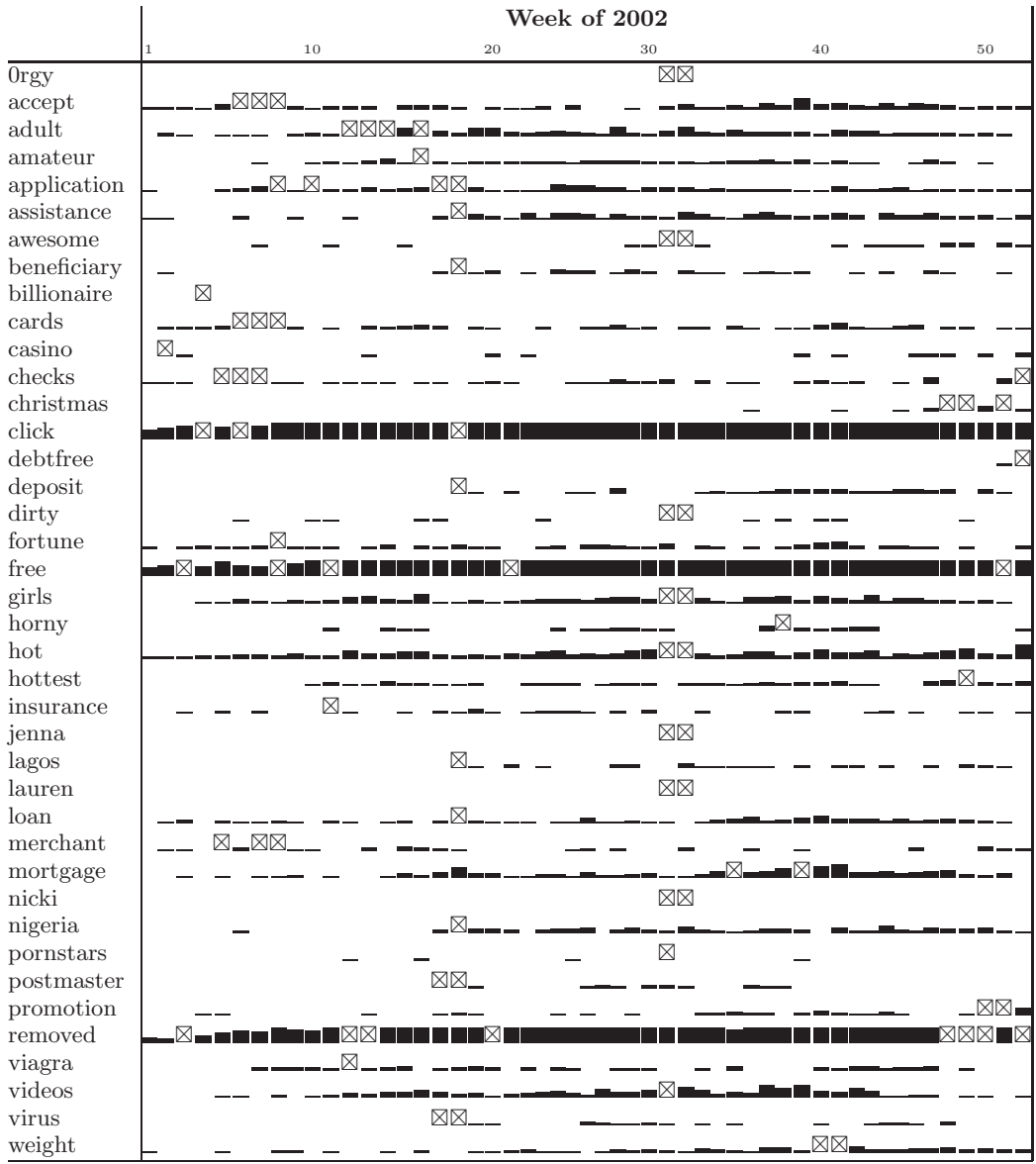

    \centering \begin{tabular}{l|cccccccccccccccccccccccccccccccccccccccccccccccccccc|}
&\multicolumn{52}{c|}{\textbf{Week of 2002}}\\
&\tiny 1&&&&&&&&&\tiny 10&&&&&&&&&&\tiny 20&&&&&&&&&&\tiny 30&&&&&&&&&&\tiny 40&&&&&&&&&&\tiny 50&&\\
\hline
 0rgy~~ & \rule{2mm}{0mm}&\rule{2mm}{0mm}&\rule{2mm}{0mm}&\rule{2mm}{0mm}&\rule{2mm}{0mm}&\rule{2mm}{0mm}&\rule{2mm}{0mm}&\rule{2mm}{0mm}&\rule{2mm}{0mm}&\rule{2mm}{0mm}&\rule{2mm}{0mm}&\rule{2mm}{0mm}&\rule{2mm}{0mm}&\rule{2mm}{0mm}&\rule{2mm}{0mm}&\rule{2mm}{0mm}&\rule{2mm}{0mm}&\rule{2mm}{0mm}&\rule{2mm}{0mm}&\rule{2mm}{0mm}&\rule{2mm}{0mm}&\rule{2mm}{0mm}&\rule{2mm}{0mm}&\rule{2mm}{0mm}&\rule{2mm}{0mm}&\rule{2mm}{0mm}&\rule{2mm}{0mm}&\rule{2mm}{0mm}&\rule{2mm}{0mm}&\rule{2mm}{0mm}&\burst&\burst&\rule{2mm}{0mm}&\rule{2mm}{0mm}&\rule{2mm}{0mm}&\rule{2mm}{0mm}&\rule{2mm}{0mm}&\rule{2mm}{0mm}&\rule{2mm}{0mm}&\rule{2mm}{0mm}&\rule{2mm}{0mm}&\rule{2mm}{0mm}&\rule{2mm}{0mm}&\rule{2mm}{0mm}&\rule{2mm}{0mm}&\rule{2mm}{0mm}&\rule{2mm}{0mm}&\rule{2mm}{0mm}&\rule{2mm}{0mm}&\rule{2mm}{0mm}&\rule{2mm}{0mm}&\rule{2mm}{0mm}\\
 accept~~ & \rule{2mm}{0.2mm}&\rule{2mm}{0.2mm}&\rule{2mm}{0.266666666666667mm}&\rule{2mm}{0.133333333333333mm}&\rule{2mm}{0.6mm}&\burst&\burst&\burst&\rule{2mm}{0.333333333333333mm}&\rule{2mm}{0.133333333333333mm}&\rule{2mm}{0.333333333333333mm}&\rule{2mm}{0.333333333333333mm}&\rule{2mm}{0.4mm}&\rule{2mm}{0mm}&\rule{2mm}{0.466666666666667mm}&\rule{2mm}{0.466666666666667mm}&\rule{2mm}{0.466666666666667mm}&\rule{2mm}{0.266666666666667mm}&\rule{2mm}{0mm}&\rule{2mm}{0.2mm}&\rule{2mm}{0.133333333333333mm}&\rule{2mm}{0.133333333333333mm}&\rule{2mm}{0.4mm}&\rule{2mm}{0mm}&\rule{2mm}{0.533333333333333mm}&\rule{2mm}{0mm}&\rule{2mm}{0mm}&\rule{2mm}{0mm}&\rule{2mm}{0.133333333333333mm}&\rule{2mm}{0mm}&\rule{2mm}{0.4mm}&\rule{2mm}{0.6mm}&\rule{2mm}{0.266666666666667mm}&\rule{2mm}{0.266666666666667mm}&\rule{2mm}{0.533333333333333mm}&\rule{2mm}{0.266666666666667mm}&\rule{2mm}{0.733333333333333mm}&\rule{2mm}{0.466666666666667mm}&\rule{2mm}{1.4mm}&\rule{2mm}{0.6mm}&\rule{2mm}{0.8mm}&\rule{2mm}{0.466666666666667mm}&\rule{2mm}{0.4mm}&\rule{2mm}{0.733333333333333mm}&\rule{2mm}{0.4mm}&\rule{2mm}{0.8mm}&\rule{2mm}{0.6mm}&\rule{2mm}{0.466666666666667mm}&\rule{2mm}{0.266666666666667mm}&\rule{2mm}{0.333333333333333mm}&\rule{2mm}{0.4mm}&\rule{2mm}{0.4mm}\\
 adult~~ & \rule{2mm}{0mm}&\rule{2mm}{0.4mm}&\rule{2mm}{0.2mm}&\rule{2mm}{0mm}&\rule{2mm}{0.133333333333333mm}&\rule{2mm}{0.133333333333333mm}&\rule{2mm}{0.133333333333333mm}&\rule{2mm}{0mm}&\rule{2mm}{0.333333333333333mm}&\rule{2mm}{0.4mm}&\rule{2mm}{0.333333333333333mm}&\burst&\burst&\burst&\rule{2mm}{1.13333333333333mm}&\burst&\rule{2mm}{0.666666666666667mm}&\rule{2mm}{0.4mm}&\rule{2mm}{1.06666666666667mm}&\rule{2mm}{1.06666666666667mm}&\rule{2mm}{0.533333333333333mm}&\rule{2mm}{0.466666666666667mm}&\rule{2mm}{0.6mm}&\rule{2mm}{0.666666666666667mm}&\rule{2mm}{0.6mm}&\rule{2mm}{0.466666666666667mm}&\rule{2mm}{0.333333333333333mm}&\rule{2mm}{1.26666666666667mm}&\rule{2mm}{0.4mm}&\rule{2mm}{0.333333333333333mm}&\rule{2mm}{0.733333333333333mm}&\rule{2mm}{1.2mm}&\rule{2mm}{0.533333333333333mm}&\rule{2mm}{0.466666666666667mm}&\rule{2mm}{0.8mm}&\rule{2mm}{0.6mm}&\rule{2mm}{0.533333333333333mm}&\rule{2mm}{0.533333333333333mm}&\rule{2mm}{0.6mm}&\rule{2mm}{0.266666666666667mm}&\rule{2mm}{0.866666666666667mm}&\rule{2mm}{0.733333333333333mm}&\rule{2mm}{0.733333333333333mm}&\rule{2mm}{0.266666666666667mm}&\rule{2mm}{0.466666666666667mm}&\rule{2mm}{0.466666666666667mm}&\rule{2mm}{0.466666666666667mm}&\rule{2mm}{0.466666666666667mm}&\rule{2mm}{0.333333333333333mm}&\rule{2mm}{0.4mm}&\rule{2mm}{0.266666666666667mm}&\rule{2mm}{0mm}\\
 amateur~~ & \rule{2mm}{0mm}&\rule{2mm}{0mm}&\rule{2mm}{0mm}&\rule{2mm}{0mm}&\rule{2mm}{0mm}&\rule{2mm}{0mm}&\rule{2mm}{0.2mm}&\rule{2mm}{0mm}&\rule{2mm}{0mm}&\rule{2mm}{0.133333333333333mm}&\rule{2mm}{0.266666666666667mm}&\rule{2mm}{0.2mm}&\rule{2mm}{0.4mm}&\rule{2mm}{0.666666666666667mm}&\rule{2mm}{0.133333333333333mm}&\burst&\rule{2mm}{0.333333333333333mm}&\rule{2mm}{0.133333333333333mm}&\rule{2mm}{0.266666666666667mm}&\rule{2mm}{0.266666666666667mm}&\rule{2mm}{0.266666666666667mm}&\rule{2mm}{0.333333333333333mm}&\rule{2mm}{0.266666666666667mm}&\rule{2mm}{0.266666666666667mm}&\rule{2mm}{0.2mm}&\rule{2mm}{0.4mm}&\rule{2mm}{0.4mm}&\rule{2mm}{0.4mm}&\rule{2mm}{0.4mm}&\rule{2mm}{0.333333333333333mm}&\rule{2mm}{0.266666666666667mm}&\rule{2mm}{0.266666666666667mm}&\rule{2mm}{0.133333333333333mm}&\rule{2mm}{0.266666666666667mm}&\rule{2mm}{0.4mm}&\rule{2mm}{0.4mm}&\rule{2mm}{0.6mm}&\rule{2mm}{0.266666666666667mm}&\rule{2mm}{0.533333333333333mm}&\rule{2mm}{0.133333333333333mm}&\rule{2mm}{0.4mm}&\rule{2mm}{0.133333333333333mm}&\rule{2mm}{0.2mm}&\rule{2mm}{0mm}&\rule{2mm}{0mm}&\rule{2mm}{0.2mm}&\rule{2mm}{0.533333333333333mm}&\rule{2mm}{0.333333333333333mm}&\rule{2mm}{0mm}&\rule{2mm}{0.133333333333333mm}&\rule{2mm}{0mm}&\rule{2mm}{0mm}\\
 application~~ & \rule{2mm}{0.133333333333333mm}&\rule{2mm}{0mm}&\rule{2mm}{0mm}&\rule{2mm}{0mm}&\rule{2mm}{0.333333333333333mm}&\rule{2mm}{0.4mm}&\rule{2mm}{0.6mm}&\burst&\rule{2mm}{0.2mm}&\burst&\rule{2mm}{0.333333333333333mm}&\rule{2mm}{0.333333333333333mm}&\rule{2mm}{0.466666666666667mm}&\rule{2mm}{0.333333333333333mm}&\rule{2mm}{0.4mm}&\rule{2mm}{0.466666666666667mm}&\burst&\burst&\rule{2mm}{0.466666666666667mm}&\rule{2mm}{0.2mm}&\rule{2mm}{0.133333333333333mm}&\rule{2mm}{0.133333333333333mm}&\rule{2mm}{0.2mm}&\rule{2mm}{0.866666666666667mm}&\rule{2mm}{0.733333333333333mm}&\rule{2mm}{0.733333333333333mm}&\rule{2mm}{0.533333333333333mm}&\rule{2mm}{0.466666666666667mm}&\rule{2mm}{0.2mm}&\rule{2mm}{0.533333333333333mm}&\rule{2mm}{0.466666666666667mm}&\rule{2mm}{0.466666666666667mm}&\rule{2mm}{0.266666666666667mm}&\rule{2mm}{0.4mm}&\rule{2mm}{0.333333333333333mm}&\rule{2mm}{0.333333333333333mm}&\rule{2mm}{0.266666666666667mm}&\rule{2mm}{0.333333333333333mm}&\rule{2mm}{0.2mm}&\rule{2mm}{0.133333333333333mm}&\rule{2mm}{0.666666666666667mm}&\rule{2mm}{0.266666666666667mm}&\rule{2mm}{0.333333333333333mm}&\rule{2mm}{0.4mm}&\rule{2mm}{0.533333333333333mm}&\rule{2mm}{0.2mm}&\rule{2mm}{0.266666666666667mm}&\rule{2mm}{0.333333333333333mm}&\rule{2mm}{0.266666666666667mm}&\rule{2mm}{0.333333333333333mm}&\rule{2mm}{0.266666666666667mm}&\rule{2mm}{0.333333333333333mm}\\
 assistance~~ & \rule{2mm}{0.133333333333333mm}&\rule{2mm}{0.133333333333333mm}&\rule{2mm}{0mm}&\rule{2mm}{0mm}&\rule{2mm}{0mm}&\rule{2mm}{0.333333333333333mm}&\rule{2mm}{0mm}&\rule{2mm}{0mm}&\rule{2mm}{0.2mm}&\rule{2mm}{0mm}&\rule{2mm}{0mm}&\rule{2mm}{0.2mm}&\rule{2mm}{0mm}&\rule{2mm}{0mm}&\rule{2mm}{0mm}&\rule{2mm}{0mm}&\rule{2mm}{0.333333333333333mm}&\burst&\rule{2mm}{0.6mm}&\rule{2mm}{0.4mm}&\rule{2mm}{0.133333333333333mm}&\rule{2mm}{0.733333333333333mm}&\rule{2mm}{0.133333333333333mm}&\rule{2mm}{0.733333333333333mm}&\rule{2mm}{0.8mm}&\rule{2mm}{0.666666666666667mm}&\rule{2mm}{0.133333333333333mm}&\rule{2mm}{0.6mm}&\rule{2mm}{0.333333333333333mm}&\rule{2mm}{0.4mm}&\rule{2mm}{0.2mm}&\rule{2mm}{0.933333333333333mm}&\rule{2mm}{0.6mm}&\rule{2mm}{0.2mm}&\rule{2mm}{0.133333333333333mm}&\rule{2mm}{0.6mm}&\rule{2mm}{0.866666666666667mm}&\rule{2mm}{0.533333333333333mm}&\rule{2mm}{0.333333333333333mm}&\rule{2mm}{0.533333333333333mm}&\rule{2mm}{0.8mm}&\rule{2mm}{0.466666666666667mm}&\rule{2mm}{0mm}&\rule{2mm}{0.8mm}&\rule{2mm}{0.466666666666667mm}&\rule{2mm}{0.466666666666667mm}&\rule{2mm}{0.733333333333333mm}&\rule{2mm}{0.4mm}&\rule{2mm}{0.4mm}&\rule{2mm}{0.533333333333333mm}&\rule{2mm}{0.133333333333333mm}&\rule{2mm}{0.533333333333333mm}\\
 awesome~~ & \rule{2mm}{0mm}&\rule{2mm}{0mm}&\rule{2mm}{0mm}&\rule{2mm}{0mm}&\rule{2mm}{0mm}&\rule{2mm}{0mm}&\rule{2mm}{0.133333333333333mm}&\rule{2mm}{0mm}&\rule{2mm}{0mm}&\rule{2mm}{0mm}&\rule{2mm}{0.133333333333333mm}&\rule{2mm}{0mm}&\rule{2mm}{0mm}&\rule{2mm}{0mm}&\rule{2mm}{0.133333333333333mm}&\rule{2mm}{0mm}&\rule{2mm}{0mm}&\rule{2mm}{0mm}&\rule{2mm}{0mm}&\rule{2mm}{0mm}&\rule{2mm}{0mm}&\rule{2mm}{0mm}&\rule{2mm}{0mm}&\rule{2mm}{0mm}&\rule{2mm}{0mm}&\rule{2mm}{0mm}&\rule{2mm}{0mm}&\rule{2mm}{0mm}&\rule{2mm}{0.133333333333333mm}&\rule{2mm}{0.133333333333333mm}&\burst&\burst&\rule{2mm}{0.133333333333333mm}&\rule{2mm}{0mm}&\rule{2mm}{0mm}&\rule{2mm}{0mm}&\rule{2mm}{0mm}&\rule{2mm}{0mm}&\rule{2mm}{0mm}&\rule{2mm}{0mm}&\rule{2mm}{0.2mm}&\rule{2mm}{0mm}&\rule{2mm}{0.2mm}&\rule{2mm}{0.133333333333333mm}&\rule{2mm}{0.133333333333333mm}&\rule{2mm}{0.2mm}&\rule{2mm}{0mm}&\rule{2mm}{0.4mm}&\rule{2mm}{0.4mm}&\rule{2mm}{0mm}&\rule{2mm}{0.4mm}&\rule{2mm}{0.133333333333333mm}\\
 beneficiary~~ & \rule{2mm}{0mm}&\rule{2mm}{0.133333333333333mm}&\rule{2mm}{0mm}&\rule{2mm}{0mm}&\rule{2mm}{0mm}&\rule{2mm}{0mm}&\rule{2mm}{0mm}&\rule{2mm}{0mm}&\rule{2mm}{0mm}&\rule{2mm}{0mm}&\rule{2mm}{0mm}&\rule{2mm}{0mm}&\rule{2mm}{0mm}&\rule{2mm}{0mm}&\rule{2mm}{0mm}&\rule{2mm}{0mm}&\rule{2mm}{0.333333333333333mm}&\burst&\rule{2mm}{0.2mm}&\rule{2mm}{0.4mm}&\rule{2mm}{0mm}&\rule{2mm}{0.2mm}&\rule{2mm}{0mm}&\rule{2mm}{0.466666666666667mm}&\rule{2mm}{0.4mm}&\rule{2mm}{0.4mm}&\rule{2mm}{0mm}&\rule{2mm}{0.2mm}&\rule{2mm}{0.466666666666667mm}&\rule{2mm}{0.266666666666667mm}&\rule{2mm}{0mm}&\rule{2mm}{0.4mm}&\rule{2mm}{0.2mm}&\rule{2mm}{0.133333333333333mm}&\rule{2mm}{0mm}&\rule{2mm}{0.2mm}&\rule{2mm}{0.266666666666667mm}&\rule{2mm}{0.133333333333333mm}&\rule{2mm}{0.266666666666667mm}&\rule{2mm}{0mm}&\rule{2mm}{0mm}&\rule{2mm}{0.133333333333333mm}&\rule{2mm}{0mm}&\rule{2mm}{0.266666666666667mm}&\rule{2mm}{0mm}&\rule{2mm}{0mm}&\rule{2mm}{0.4mm}&\rule{2mm}{0.133333333333333mm}&\rule{2mm}{0mm}&\rule{2mm}{0mm}&\rule{2mm}{0.133333333333333mm}&\rule{2mm}{0mm}\\
 billionaire~~ & \rule{2mm}{0mm}&\rule{2mm}{0mm}&\rule{2mm}{0mm}&\burst&\rule{2mm}{0mm}&\rule{2mm}{0mm}&\rule{2mm}{0mm}&\rule{2mm}{0mm}&\rule{2mm}{0mm}&\rule{2mm}{0mm}&\rule{2mm}{0mm}&\rule{2mm}{0mm}&\rule{2mm}{0mm}&\rule{2mm}{0mm}&\rule{2mm}{0mm}&\rule{2mm}{0mm}&\rule{2mm}{0mm}&\rule{2mm}{0mm}&\rule{2mm}{0mm}&\rule{2mm}{0mm}&\rule{2mm}{0mm}&\rule{2mm}{0mm}&\rule{2mm}{0mm}&\rule{2mm}{0mm}&\rule{2mm}{0mm}&\rule{2mm}{0mm}&\rule{2mm}{0mm}&\rule{2mm}{0mm}&\rule{2mm}{0mm}&\rule{2mm}{0mm}&\rule{2mm}{0mm}&\rule{2mm}{0mm}&\rule{2mm}{0mm}&\rule{2mm}{0mm}&\rule{2mm}{0mm}&\rule{2mm}{0mm}&\rule{2mm}{0mm}&\rule{2mm}{0mm}&\rule{2mm}{0mm}&\rule{2mm}{0mm}&\rule{2mm}{0mm}&\rule{2mm}{0mm}&\rule{2mm}{0mm}&\rule{2mm}{0mm}&\rule{2mm}{0mm}&\rule{2mm}{0mm}&\rule{2mm}{0mm}&\rule{2mm}{0mm}&\rule{2mm}{0mm}&\rule{2mm}{0mm}&\rule{2mm}{0mm}&\rule{2mm}{0mm}\\
 cards~~ & \rule{2mm}{0mm}&\rule{2mm}{0.266666666666667mm}&\rule{2mm}{0.2mm}&\rule{2mm}{0.2mm}&\rule{2mm}{0.4mm}&\burst&\burst&\burst&\rule{2mm}{0.266666666666667mm}&\rule{2mm}{0mm}&\rule{2mm}{0.133333333333333mm}&\rule{2mm}{0mm}&\rule{2mm}{0.4mm}&\rule{2mm}{0.2mm}&\rule{2mm}{0.4mm}&\rule{2mm}{0.466666666666667mm}&\rule{2mm}{0.333333333333333mm}&\rule{2mm}{0mm}&\rule{2mm}{0.266666666666667mm}&\rule{2mm}{0.133333333333333mm}&\rule{2mm}{0mm}&\rule{2mm}{0mm}&\rule{2mm}{0.2mm}&\rule{2mm}{0mm}&\rule{2mm}{0mm}&\rule{2mm}{0.266666666666667mm}&\rule{2mm}{0.266666666666667mm}&\rule{2mm}{0.533333333333333mm}&\rule{2mm}{0.133333333333333mm}&\rule{2mm}{0.2mm}&\rule{2mm}{0mm}&\rule{2mm}{0.266666666666667mm}&\rule{2mm}{0.2mm}&\rule{2mm}{0mm}&\rule{2mm}{0.4mm}&\rule{2mm}{0.133333333333333mm}&\rule{2mm}{0mm}&\rule{2mm}{0.133333333333333mm}&\rule{2mm}{0.133333333333333mm}&\rule{2mm}{0.466666666666667mm}&\rule{2mm}{0.733333333333333mm}&\rule{2mm}{0.266666666666667mm}&\rule{2mm}{0.133333333333333mm}&\rule{2mm}{0.133333333333333mm}&\rule{2mm}{0.333333333333333mm}&\rule{2mm}{0.466666666666667mm}&\rule{2mm}{0mm}&\rule{2mm}{0.2mm}&\rule{2mm}{0.266666666666667mm}&\rule{2mm}{0mm}&\rule{2mm}{0.133333333333333mm}&\rule{2mm}{0.133333333333333mm}\\
 casino~~ & \rule{2mm}{0mm}&\burst&\rule{2mm}{0.133333333333333mm}&\rule{2mm}{0mm}&\rule{2mm}{0mm}&\rule{2mm}{0mm}&\rule{2mm}{0mm}&\rule{2mm}{0mm}&\rule{2mm}{0mm}&\rule{2mm}{0mm}&\rule{2mm}{0mm}&\rule{2mm}{0mm}&\rule{2mm}{0.133333333333333mm}&\rule{2mm}{0mm}&\rule{2mm}{0mm}&\rule{2mm}{0mm}&\rule{2mm}{0mm}&\rule{2mm}{0mm}&\rule{2mm}{0mm}&\rule{2mm}{0.266666666666667mm}&\rule{2mm}{0mm}&\rule{2mm}{0.133333333333333mm}&\rule{2mm}{0mm}&\rule{2mm}{0mm}&\rule{2mm}{0mm}&\rule{2mm}{0mm}&\rule{2mm}{0mm}&\rule{2mm}{0mm}&\rule{2mm}{0mm}&\rule{2mm}{0mm}&\rule{2mm}{0mm}&\rule{2mm}{0mm}&\rule{2mm}{0mm}&\rule{2mm}{0mm}&\rule{2mm}{0mm}&\rule{2mm}{0mm}&\rule{2mm}{0mm}&\rule{2mm}{0mm}&\rule{2mm}{0.333333333333333mm}&\rule{2mm}{0mm}&\rule{2mm}{0.133333333333333mm}&\rule{2mm}{0mm}&\rule{2mm}{0mm}&\rule{2mm}{0mm}&\rule{2mm}{0mm}&\rule{2mm}{0.266666666666667mm}&\rule{2mm}{0.333333333333333mm}&\rule{2mm}{0.333333333333333mm}&\rule{2mm}{0mm}&\rule{2mm}{0.266666666666667mm}&\rule{2mm}{0mm}&\rule{2mm}{0.4mm}\\
 checks~~ & \rule{2mm}{0.133333333333333mm}&\rule{2mm}{0.133333333333333mm}&\rule{2mm}{0.2mm}&\rule{2mm}{0mm}&\burst&\burst&\burst&\rule{2mm}{0.2mm}&\rule{2mm}{0.2mm}&\rule{2mm}{0mm}&\rule{2mm}{0.2mm}&\rule{2mm}{0.133333333333333mm}&\rule{2mm}{0.2mm}&\rule{2mm}{0.133333333333333mm}&\rule{2mm}{0mm}&\rule{2mm}{0.133333333333333mm}&\rule{2mm}{0.133333333333333mm}&\rule{2mm}{0mm}&\rule{2mm}{0.133333333333333mm}&\rule{2mm}{0.333333333333333mm}&\rule{2mm}{0.133333333333333mm}&\rule{2mm}{0mm}&\rule{2mm}{0mm}&\rule{2mm}{0mm}&\rule{2mm}{0.2mm}&\rule{2mm}{0.2mm}&\rule{2mm}{0.2mm}&\rule{2mm}{0.466666666666667mm}&\rule{2mm}{0.266666666666667mm}&\rule{2mm}{0.266666666666667mm}&\rule{2mm}{0.466666666666667mm}&\rule{2mm}{0mm}&\rule{2mm}{0.4mm}&\rule{2mm}{0mm}&\rule{2mm}{0.133333333333333mm}&\rule{2mm}{0.133333333333333mm}&\rule{2mm}{0mm}&\rule{2mm}{0mm}&\rule{2mm}{0.333333333333333mm}&\rule{2mm}{0.4mm}&\rule{2mm}{0.2mm}&\rule{2mm}{0.266666666666667mm}&\rule{2mm}{0mm}&\rule{2mm}{0mm}&\rule{2mm}{0.2mm}&\rule{2mm}{0mm}&\rule{2mm}{0.8mm}&\rule{2mm}{0mm}&\rule{2mm}{0mm}&\rule{2mm}{0mm}&\rule{2mm}{0.6mm}&\burst\\
 christmas~~ & \rule{2mm}{0mm}&\rule{2mm}{0mm}&\rule{2mm}{0mm}&\rule{2mm}{0mm}&\rule{2mm}{0mm}&\rule{2mm}{0mm}&\rule{2mm}{0mm}&\rule{2mm}{0mm}&\rule{2mm}{0mm}&\rule{2mm}{0mm}&\rule{2mm}{0mm}&\rule{2mm}{0mm}&\rule{2mm}{0mm}&\rule{2mm}{0mm}&\rule{2mm}{0mm}&\rule{2mm}{0mm}&\rule{2mm}{0mm}&\rule{2mm}{0mm}&\rule{2mm}{0mm}&\rule{2mm}{0mm}&\rule{2mm}{0mm}&\rule{2mm}{0mm}&\rule{2mm}{0mm}&\rule{2mm}{0mm}&\rule{2mm}{0mm}&\rule{2mm}{0mm}&\rule{2mm}{0mm}&\rule{2mm}{0mm}&\rule{2mm}{0mm}&\rule{2mm}{0mm}&\rule{2mm}{0mm}&\rule{2mm}{0mm}&\rule{2mm}{0mm}&\rule{2mm}{0mm}&\rule{2mm}{0mm}&\rule{2mm}{0.133333333333333mm}&\rule{2mm}{0mm}&\rule{2mm}{0mm}&\rule{2mm}{0mm}&\rule{2mm}{0mm}&\rule{2mm}{0.133333333333333mm}&\rule{2mm}{0mm}&\rule{2mm}{0mm}&\rule{2mm}{0mm}&\rule{2mm}{0.133333333333333mm}&\rule{2mm}{0mm}&\rule{2mm}{0.333333333333333mm}&\burst&\burst&\rule{2mm}{0.666666666666667mm}&\burst&\rule{2mm}{0.2mm}\\
 click~~ & \rule{2mm}{1.06666666666667mm}&\rule{2mm}{1.33333333333333mm}&\rule{2mm}{1.73333333333333mm}&\burst&\rule{2mm}{1.6mm}&\burst&\rule{2mm}{1.66666666666667mm}&\rule{2mm}{2mm}&\rule{2mm}{2mm}&\rule{2mm}{2mm}&\rule{2mm}{2mm}&\rule{2mm}{2mm}&\rule{2mm}{2mm}&\rule{2mm}{2mm}&\rule{2mm}{2mm}&\rule{2mm}{2mm}&\rule{2mm}{2mm}&\burst&\rule{2mm}{2mm}&\rule{2mm}{2mm}&\rule{2mm}{2mm}&\rule{2mm}{2mm}&\rule{2mm}{2mm}&\rule{2mm}{2mm}&\rule{2mm}{2mm}&\rule{2mm}{2mm}&\rule{2mm}{2mm}&\rule{2mm}{2mm}&\rule{2mm}{2mm}&\rule{2mm}{2mm}&\rule{2mm}{2mm}&\rule{2mm}{2mm}&\rule{2mm}{2mm}&\rule{2mm}{2mm}&\rule{2mm}{2mm}&\rule{2mm}{2mm}&\rule{2mm}{2mm}&\rule{2mm}{2mm}&\rule{2mm}{2mm}&\rule{2mm}{2mm}&\rule{2mm}{2mm}&\rule{2mm}{2mm}&\rule{2mm}{2mm}&\rule{2mm}{2mm}&\rule{2mm}{2mm}&\rule{2mm}{2mm}&\rule{2mm}{2mm}&\rule{2mm}{2mm}&\rule{2mm}{2mm}&\rule{2mm}{2mm}&\rule{2mm}{2mm}&\rule{2mm}{2mm}\\
 debtfree~~ & \rule{2mm}{0mm}&\rule{2mm}{0mm}&\rule{2mm}{0mm}&\rule{2mm}{0mm}&\rule{2mm}{0mm}&\rule{2mm}{0mm}&\rule{2mm}{0mm}&\rule{2mm}{0mm}&\rule{2mm}{0mm}&\rule{2mm}{0mm}&\rule{2mm}{0mm}&\rule{2mm}{0mm}&\rule{2mm}{0mm}&\rule{2mm}{0mm}&\rule{2mm}{0mm}&\rule{2mm}{0mm}&\rule{2mm}{0mm}&\rule{2mm}{0mm}&\rule{2mm}{0mm}&\rule{2mm}{0mm}&\rule{2mm}{0mm}&\rule{2mm}{0mm}&\rule{2mm}{0mm}&\rule{2mm}{0mm}&\rule{2mm}{0mm}&\rule{2mm}{0mm}&\rule{2mm}{0mm}&\rule{2mm}{0mm}&\rule{2mm}{0mm}&\rule{2mm}{0mm}&\rule{2mm}{0mm}&\rule{2mm}{0mm}&\rule{2mm}{0mm}&\rule{2mm}{0mm}&\rule{2mm}{0mm}&\rule{2mm}{0mm}&\rule{2mm}{0mm}&\rule{2mm}{0mm}&\rule{2mm}{0mm}&\rule{2mm}{0mm}&\rule{2mm}{0mm}&\rule{2mm}{0mm}&\rule{2mm}{0mm}&\rule{2mm}{0mm}&\rule{2mm}{0mm}&\rule{2mm}{0mm}&\rule{2mm}{0mm}&\rule{2mm}{0mm}&\rule{2mm}{0mm}&\rule{2mm}{0mm}&\rule{2mm}{0.333333333333333mm}&\burst\\
 deposit~~ & \rule{2mm}{0mm}&\rule{2mm}{0mm}&\rule{2mm}{0mm}&\rule{2mm}{0mm}&\rule{2mm}{0mm}&\rule{2mm}{0mm}&\rule{2mm}{0mm}&\rule{2mm}{0mm}&\rule{2mm}{0mm}&\rule{2mm}{0mm}&\rule{2mm}{0mm}&\rule{2mm}{0mm}&\rule{2mm}{0mm}&\rule{2mm}{0mm}&\rule{2mm}{0mm}&\rule{2mm}{0mm}&\rule{2mm}{0mm}&\burst&\rule{2mm}{0.2mm}&\rule{2mm}{0mm}&\rule{2mm}{0.333333333333333mm}&\rule{2mm}{0mm}&\rule{2mm}{0mm}&\rule{2mm}{0mm}&\rule{2mm}{0.133333333333333mm}&\rule{2mm}{0.2mm}&\rule{2mm}{0mm}&\rule{2mm}{0.733333333333333mm}&\rule{2mm}{0mm}&\rule{2mm}{0mm}&\rule{2mm}{0mm}&\rule{2mm}{0mm}&\rule{2mm}{0.2mm}&\rule{2mm}{0.333333333333333mm}&\rule{2mm}{0.133333333333333mm}&\rule{2mm}{0.133333333333333mm}&\rule{2mm}{0.333333333333333mm}&\rule{2mm}{0.533333333333333mm}&\rule{2mm}{0.6mm}&\rule{2mm}{0.533333333333333mm}&\rule{2mm}{0.466666666666667mm}&\rule{2mm}{0.266666666666667mm}&\rule{2mm}{0.266666666666667mm}&\rule{2mm}{0.333333333333333mm}&\rule{2mm}{0.466666666666667mm}&\rule{2mm}{0.6mm}&\rule{2mm}{0.4mm}&\rule{2mm}{0.533333333333333mm}&\rule{2mm}{0mm}&\rule{2mm}{0.466666666666667mm}&\rule{2mm}{0.2mm}&\rule{2mm}{0mm}\\
 dirty~~ & \rule{2mm}{0mm}&\rule{2mm}{0mm}&\rule{2mm}{0mm}&\rule{2mm}{0mm}&\rule{2mm}{0mm}&\rule{2mm}{0.133333333333333mm}&\rule{2mm}{0mm}&\rule{2mm}{0mm}&\rule{2mm}{0mm}&\rule{2mm}{0.133333333333333mm}&\rule{2mm}{0.133333333333333mm}&\rule{2mm}{0mm}&\rule{2mm}{0mm}&\rule{2mm}{0mm}&\rule{2mm}{0mm}&\rule{2mm}{0.2mm}&\rule{2mm}{0.2mm}&\rule{2mm}{0mm}&\rule{2mm}{0mm}&\rule{2mm}{0mm}&\rule{2mm}{0mm}&\rule{2mm}{0mm}&\rule{2mm}{0.2mm}&\rule{2mm}{0mm}&\rule{2mm}{0mm}&\rule{2mm}{0mm}&\rule{2mm}{0mm}&\rule{2mm}{0mm}&\rule{2mm}{0mm}&\rule{2mm}{0mm}&\burst&\burst&\rule{2mm}{0mm}&\rule{2mm}{0mm}&\rule{2mm}{0mm}&\rule{2mm}{0.133333333333333mm}&\rule{2mm}{0mm}&\rule{2mm}{0.2mm}&\rule{2mm}{0mm}&\rule{2mm}{0.2mm}&\rule{2mm}{0.2mm}&\rule{2mm}{0mm}&\rule{2mm}{0mm}&\rule{2mm}{0mm}&\rule{2mm}{0mm}&\rule{2mm}{0mm}&\rule{2mm}{0mm}&\rule{2mm}{0mm}&\rule{2mm}{0.133333333333333mm}&\rule{2mm}{0mm}&\rule{2mm}{0mm}&\rule{2mm}{0mm}\\
 fortune~~ & \rule{2mm}{0.2mm}&\rule{2mm}{0mm}&\rule{2mm}{0.133333333333333mm}&\rule{2mm}{0.333333333333333mm}&\rule{2mm}{0.2mm}&\rule{2mm}{0.133333333333333mm}&\rule{2mm}{0.266666666666667mm}&\burst&\rule{2mm}{0.133333333333333mm}&\rule{2mm}{0.2mm}&\rule{2mm}{0.133333333333333mm}&\rule{2mm}{0mm}&\rule{2mm}{0.133333333333333mm}&\rule{2mm}{0.466666666666667mm}&\rule{2mm}{0mm}&\rule{2mm}{0.266666666666667mm}&\rule{2mm}{0.133333333333333mm}&\rule{2mm}{0.4mm}&\rule{2mm}{0.2mm}&\rule{2mm}{0.2mm}&\rule{2mm}{0mm}&\rule{2mm}{0mm}&\rule{2mm}{0.2mm}&\rule{2mm}{0.266666666666667mm}&\rule{2mm}{0mm}&\rule{2mm}{0.466666666666667mm}&\rule{2mm}{0.533333333333333mm}&\rule{2mm}{0.266666666666667mm}&\rule{2mm}{0.2mm}&\rule{2mm}{0.2mm}&\rule{2mm}{0.666666666666667mm}&\rule{2mm}{0mm}&\rule{2mm}{0.266666666666667mm}&\rule{2mm}{0.2mm}&\rule{2mm}{0mm}&\rule{2mm}{0.133333333333333mm}&\rule{2mm}{0mm}&\rule{2mm}{0.133333333333333mm}&\rule{2mm}{0.4mm}&\rule{2mm}{0.733333333333333mm}&\rule{2mm}{0.866666666666667mm}&\rule{2mm}{0.333333333333333mm}&\rule{2mm}{0mm}&\rule{2mm}{0.333333333333333mm}&\rule{2mm}{0.333333333333333mm}&\rule{2mm}{0.133333333333333mm}&\rule{2mm}{0.2mm}&\rule{2mm}{0mm}&\rule{2mm}{0.2mm}&\rule{2mm}{0mm}&\rule{2mm}{0mm}&\rule{2mm}{0.266666666666667mm}\\
 free~~ & \rule{2mm}{1.13333333333333mm}&\rule{2mm}{1.33333333333333mm}&\burst&\rule{2mm}{1.26666666666667mm}&\rule{2mm}{1.93333333333333mm}&\rule{2mm}{1.4mm}&\rule{2mm}{1.26666666666667mm}&\burst&\rule{2mm}{1.6mm}&\rule{2mm}{2mm}&\burst&\rule{2mm}{2mm}&\rule{2mm}{2mm}&\rule{2mm}{2mm}&\rule{2mm}{2mm}&\rule{2mm}{2mm}&\rule{2mm}{2mm}&\rule{2mm}{2mm}&\rule{2mm}{2mm}&\rule{2mm}{2mm}&\burst&\rule{2mm}{2mm}&\rule{2mm}{2mm}&\rule{2mm}{2mm}&\rule{2mm}{2mm}&\rule{2mm}{2mm}&\rule{2mm}{2mm}&\rule{2mm}{2mm}&\rule{2mm}{2mm}&\rule{2mm}{2mm}&\rule{2mm}{2mm}&\rule{2mm}{2mm}&\rule{2mm}{2mm}&\rule{2mm}{2mm}&\rule{2mm}{2mm}&\rule{2mm}{2mm}&\rule{2mm}{2mm}&\rule{2mm}{2mm}&\rule{2mm}{2mm}&\rule{2mm}{2mm}&\rule{2mm}{2mm}&\rule{2mm}{2mm}&\rule{2mm}{2mm}&\rule{2mm}{2mm}&\rule{2mm}{2mm}&\rule{2mm}{2mm}&\rule{2mm}{2mm}&\rule{2mm}{2mm}&\rule{2mm}{2mm}&\rule{2mm}{2mm}&\burst&\rule{2mm}{2mm}\\
 girls~~ & \rule{2mm}{0mm}&\rule{2mm}{0mm}&\rule{2mm}{0mm}&\rule{2mm}{0.133333333333333mm}&\rule{2mm}{0.133333333333333mm}&\rule{2mm}{0.466666666666667mm}&\rule{2mm}{0.2mm}&\rule{2mm}{0.133333333333333mm}&\rule{2mm}{0.4mm}&\rule{2mm}{0.266666666666667mm}&\rule{2mm}{0.4mm}&\rule{2mm}{0.8mm}&\rule{2mm}{0.866666666666667mm}&\rule{2mm}{0.533333333333333mm}&\rule{2mm}{0.4mm}&\rule{2mm}{1.13333333333333mm}&\rule{2mm}{0.133333333333333mm}&\rule{2mm}{0.133333333333333mm}&\rule{2mm}{0.333333333333333mm}&\rule{2mm}{0.133333333333333mm}&\rule{2mm}{0.266666666666667mm}&\rule{2mm}{0.333333333333333mm}&\rule{2mm}{0.533333333333333mm}&\rule{2mm}{0.466666666666667mm}&\rule{2mm}{0.533333333333333mm}&\rule{2mm}{0.4mm}&\rule{2mm}{0.6mm}&\rule{2mm}{0.8mm}&\rule{2mm}{0.733333333333333mm}&\rule{2mm}{0.466666666666667mm}&\burst&\burst&\rule{2mm}{0.6mm}&\rule{2mm}{0.2mm}&\rule{2mm}{0.133333333333333mm}&\rule{2mm}{0.733333333333333mm}&\rule{2mm}{0.733333333333333mm}&\rule{2mm}{0.933333333333333mm}&\rule{2mm}{0.333333333333333mm}&\rule{2mm}{0.933333333333333mm}&\rule{2mm}{0.666666666666667mm}&\rule{2mm}{0.4mm}&\rule{2mm}{1mm}&\rule{2mm}{0.266666666666667mm}&\rule{2mm}{0.666666666666667mm}&\rule{2mm}{0.6mm}&\rule{2mm}{0.333333333333333mm}&\rule{2mm}{0.4mm}&\rule{2mm}{0.266666666666667mm}&\rule{2mm}{0.333333333333333mm}&\rule{2mm}{0.133333333333333mm}&\rule{2mm}{0mm}\\
 horny~~ & \rule{2mm}{0mm}&\rule{2mm}{0mm}&\rule{2mm}{0mm}&\rule{2mm}{0mm}&\rule{2mm}{0mm}&\rule{2mm}{0mm}&\rule{2mm}{0mm}&\rule{2mm}{0mm}&\rule{2mm}{0mm}&\rule{2mm}{0mm}&\rule{2mm}{0.266666666666667mm}&\rule{2mm}{0mm}&\rule{2mm}{0mm}&\rule{2mm}{0.266666666666667mm}&\rule{2mm}{0.2mm}&\rule{2mm}{0.133333333333333mm}&\rule{2mm}{0mm}&\rule{2mm}{0mm}&\rule{2mm}{0mm}&\rule{2mm}{0mm}&\rule{2mm}{0mm}&\rule{2mm}{0mm}&\rule{2mm}{0mm}&\rule{2mm}{0.266666666666667mm}&\rule{2mm}{0mm}&\rule{2mm}{0.2mm}&\rule{2mm}{0.2mm}&\rule{2mm}{0.266666666666667mm}&\rule{2mm}{0.266666666666667mm}&\rule{2mm}{0.2mm}&\rule{2mm}{0.133333333333333mm}&\rule{2mm}{0mm}&\rule{2mm}{0mm}&\rule{2mm}{0mm}&\rule{2mm}{0mm}&\rule{2mm}{0mm}&\rule{2mm}{0.6mm}&\burst&\rule{2mm}{0.266666666666667mm}&\rule{2mm}{0.2mm}&\rule{2mm}{0.4mm}&\rule{2mm}{0.466666666666667mm}&\rule{2mm}{0.533333333333333mm}&\rule{2mm}{0mm}&\rule{2mm}{0mm}&\rule{2mm}{0mm}&\rule{2mm}{0mm}&\rule{2mm}{0mm}&\rule{2mm}{0mm}&\rule{2mm}{0mm}&\rule{2mm}{0mm}&\rule{2mm}{0.133333333333333mm}\\
 hot~~ & \rule{2mm}{0.2mm}&\rule{2mm}{0.2mm}&\rule{2mm}{0.133333333333333mm}&\rule{2mm}{0.333333333333333mm}&\rule{2mm}{0.333333333333333mm}&\rule{2mm}{0.466666666666667mm}&\rule{2mm}{0.466666666666667mm}&\rule{2mm}{0.333333333333333mm}&\rule{2mm}{0.666666666666667mm}&\rule{2mm}{0.333333333333333mm}&\rule{2mm}{0.4mm}&\rule{2mm}{1mm}&\rule{2mm}{0.6mm}&\rule{2mm}{0.666666666666667mm}&\rule{2mm}{0.866666666666667mm}&\rule{2mm}{0.866666666666667mm}&\rule{2mm}{0.533333333333333mm}&\rule{2mm}{0.333333333333333mm}&\rule{2mm}{0.466666666666667mm}&\rule{2mm}{0.4mm}&\rule{2mm}{0.6mm}&\rule{2mm}{0.466666666666667mm}&\rule{2mm}{0.866666666666667mm}&\rule{2mm}{0.933333333333333mm}&\rule{2mm}{0.533333333333333mm}&\rule{2mm}{0.6mm}&\rule{2mm}{0.466666666666667mm}&\rule{2mm}{0.6mm}&\rule{2mm}{0.933333333333333mm}&\rule{2mm}{1.06666666666667mm}&\burst&\burst&\rule{2mm}{0.6mm}&\rule{2mm}{0.4mm}&\rule{2mm}{0.466666666666667mm}&\rule{2mm}{0.866666666666667mm}&\rule{2mm}{0.866666666666667mm}&\rule{2mm}{0.4mm}&\rule{2mm}{0.733333333333333mm}&\rule{2mm}{1.13333333333333mm}&\rule{2mm}{0.8mm}&\rule{2mm}{0.8mm}&\rule{2mm}{0.933333333333333mm}&\rule{2mm}{0.133333333333333mm}&\rule{2mm}{0.666666666666667mm}&\rule{2mm}{0.466666666666667mm}&\rule{2mm}{0.8mm}&\rule{2mm}{0.933333333333333mm}&\rule{2mm}{1.2mm}&\rule{2mm}{0.6mm}&\rule{2mm}{0.466666666666667mm}&\rule{2mm}{1.8mm}\\
 hottest~~ & \rule{2mm}{0mm}&\rule{2mm}{0mm}&\rule{2mm}{0mm}&\rule{2mm}{0mm}&\rule{2mm}{0mm}&\rule{2mm}{0mm}&\rule{2mm}{0mm}&\rule{2mm}{0mm}&\rule{2mm}{0mm}&\rule{2mm}{0.133333333333333mm}&\rule{2mm}{0.4mm}&\rule{2mm}{0.2mm}&\rule{2mm}{0.133333333333333mm}&\rule{2mm}{0.533333333333333mm}&\rule{2mm}{0.266666666666667mm}&\rule{2mm}{0.266666666666667mm}&\rule{2mm}{0.133333333333333mm}&\rule{2mm}{0.2mm}&\rule{2mm}{0.266666666666667mm}&\rule{2mm}{0mm}&\rule{2mm}{0mm}&\rule{2mm}{0.266666666666667mm}&\rule{2mm}{0.266666666666667mm}&\rule{2mm}{0.333333333333333mm}&\rule{2mm}{0.333333333333333mm}&\rule{2mm}{0mm}&\rule{2mm}{0.2mm}&\rule{2mm}{0.333333333333333mm}&\rule{2mm}{0.333333333333333mm}&\rule{2mm}{0.333333333333333mm}&\rule{2mm}{0mm}&\rule{2mm}{0.266666666666667mm}&\rule{2mm}{0.333333333333333mm}&\rule{2mm}{0.266666666666667mm}&\rule{2mm}{0.133333333333333mm}&\rule{2mm}{0.333333333333333mm}&\rule{2mm}{0.4mm}&\rule{2mm}{0.333333333333333mm}&\rule{2mm}{0.466666666666667mm}&\rule{2mm}{0.466666666666667mm}&\rule{2mm}{0.6mm}&\rule{2mm}{0.2mm}&\rule{2mm}{0.133333333333333mm}&\rule{2mm}{0mm}&\rule{2mm}{0mm}&\rule{2mm}{0mm}&\rule{2mm}{0.533333333333333mm}&\rule{2mm}{0.733333333333333mm}&\burst&\rule{2mm}{0.4mm}&\rule{2mm}{0.333333333333333mm}&\rule{2mm}{0.6mm}\\
 insurance~~ & \rule{2mm}{0mm}&\rule{2mm}{0mm}&\rule{2mm}{0.133333333333333mm}&\rule{2mm}{0mm}&\rule{2mm}{0.333333333333333mm}&\rule{2mm}{0mm}&\rule{2mm}{0.333333333333333mm}&\rule{2mm}{0mm}&\rule{2mm}{0mm}&\rule{2mm}{0mm}&\burst&\rule{2mm}{0.133333333333333mm}&\rule{2mm}{0mm}&\rule{2mm}{0mm}&\rule{2mm}{0.133333333333333mm}&\rule{2mm}{0mm}&\rule{2mm}{0.333333333333333mm}&\rule{2mm}{0.133333333333333mm}&\rule{2mm}{0.466666666666667mm}&\rule{2mm}{0.133333333333333mm}&\rule{2mm}{0mm}&\rule{2mm}{0.133333333333333mm}&\rule{2mm}{0.266666666666667mm}&\rule{2mm}{0.2mm}&\rule{2mm}{0.2mm}&\rule{2mm}{0.133333333333333mm}&\rule{2mm}{0mm}&\rule{2mm}{0.2mm}&\rule{2mm}{0mm}&\rule{2mm}{0.4mm}&\rule{2mm}{0mm}&\rule{2mm}{0mm}&\rule{2mm}{0.2mm}&\rule{2mm}{0mm}&\rule{2mm}{0mm}&\rule{2mm}{0mm}&\rule{2mm}{0mm}&\rule{2mm}{0.266666666666667mm}&\rule{2mm}{0.266666666666667mm}&\rule{2mm}{0mm}&\rule{2mm}{0mm}&\rule{2mm}{0mm}&\rule{2mm}{0.133333333333333mm}&\rule{2mm}{0.333333333333333mm}&\rule{2mm}{0mm}&\rule{2mm}{0.133333333333333mm}&\rule{2mm}{0mm}&\rule{2mm}{0mm}&\rule{2mm}{0.133333333333333mm}&\rule{2mm}{0.133333333333333mm}&\rule{2mm}{0mm}&\rule{2mm}{0.133333333333333mm}\\
 jenna~~ & \rule{2mm}{0mm}&\rule{2mm}{0mm}&\rule{2mm}{0mm}&\rule{2mm}{0mm}&\rule{2mm}{0mm}&\rule{2mm}{0mm}&\rule{2mm}{0mm}&\rule{2mm}{0mm}&\rule{2mm}{0mm}&\rule{2mm}{0mm}&\rule{2mm}{0mm}&\rule{2mm}{0mm}&\rule{2mm}{0mm}&\rule{2mm}{0mm}&\rule{2mm}{0mm}&\rule{2mm}{0mm}&\rule{2mm}{0mm}&\rule{2mm}{0mm}&\rule{2mm}{0mm}&\rule{2mm}{0mm}&\rule{2mm}{0mm}&\rule{2mm}{0mm}&\rule{2mm}{0mm}&\rule{2mm}{0mm}&\rule{2mm}{0mm}&\rule{2mm}{0mm}&\rule{2mm}{0mm}&\rule{2mm}{0mm}&\rule{2mm}{0mm}&\rule{2mm}{0mm}&\burst&\burst&\rule{2mm}{0mm}&\rule{2mm}{0mm}&\rule{2mm}{0mm}&\rule{2mm}{0mm}&\rule{2mm}{0mm}&\rule{2mm}{0mm}&\rule{2mm}{0mm}&\rule{2mm}{0mm}&\rule{2mm}{0mm}&\rule{2mm}{0mm}&\rule{2mm}{0mm}&\rule{2mm}{0mm}&\rule{2mm}{0mm}&\rule{2mm}{0mm}&\rule{2mm}{0mm}&\rule{2mm}{0mm}&\rule{2mm}{0mm}&\rule{2mm}{0mm}&\rule{2mm}{0mm}&\rule{2mm}{0mm}\\
 lagos~~ & \rule{2mm}{0mm}&\rule{2mm}{0mm}&\rule{2mm}{0mm}&\rule{2mm}{0mm}&\rule{2mm}{0mm}&\rule{2mm}{0mm}&\rule{2mm}{0mm}&\rule{2mm}{0mm}&\rule{2mm}{0mm}&\rule{2mm}{0mm}&\rule{2mm}{0mm}&\rule{2mm}{0mm}&\rule{2mm}{0mm}&\rule{2mm}{0mm}&\rule{2mm}{0mm}&\rule{2mm}{0mm}&\rule{2mm}{0mm}&\burst&\rule{2mm}{0.133333333333333mm}&\rule{2mm}{0mm}&\rule{2mm}{0.4mm}&\rule{2mm}{0mm}&\rule{2mm}{0.133333333333333mm}&\rule{2mm}{0mm}&\rule{2mm}{0mm}&\rule{2mm}{0mm}&\rule{2mm}{0mm}&\rule{2mm}{0.4mm}&\rule{2mm}{0.466666666666667mm}&\rule{2mm}{0mm}&\rule{2mm}{0mm}&\rule{2mm}{0.533333333333333mm}&\rule{2mm}{0.2mm}&\rule{2mm}{0.133333333333333mm}&\rule{2mm}{0.133333333333333mm}&\rule{2mm}{0.2mm}&\rule{2mm}{0.133333333333333mm}&\rule{2mm}{0mm}&\rule{2mm}{0.333333333333333mm}&\rule{2mm}{0mm}&\rule{2mm}{0.266666666666667mm}&\rule{2mm}{0.333333333333333mm}&\rule{2mm}{0mm}&\rule{2mm}{0.333333333333333mm}&\rule{2mm}{0mm}&\rule{2mm}{0mm}&\rule{2mm}{0.266666666666667mm}&\rule{2mm}{0mm}&\rule{2mm}{0.4mm}&\rule{2mm}{0.333333333333333mm}&\rule{2mm}{0.133333333333333mm}&\rule{2mm}{0mm}\\
 lauren~~ & \rule{2mm}{0mm}&\rule{2mm}{0mm}&\rule{2mm}{0mm}&\rule{2mm}{0mm}&\rule{2mm}{0mm}&\rule{2mm}{0mm}&\rule{2mm}{0mm}&\rule{2mm}{0mm}&\rule{2mm}{0mm}&\rule{2mm}{0mm}&\rule{2mm}{0mm}&\rule{2mm}{0mm}&\rule{2mm}{0mm}&\rule{2mm}{0mm}&\rule{2mm}{0mm}&\rule{2mm}{0mm}&\rule{2mm}{0mm}&\rule{2mm}{0mm}&\rule{2mm}{0mm}&\rule{2mm}{0mm}&\rule{2mm}{0mm}&\rule{2mm}{0mm}&\rule{2mm}{0mm}&\rule{2mm}{0mm}&\rule{2mm}{0mm}&\rule{2mm}{0mm}&\rule{2mm}{0mm}&\rule{2mm}{0mm}&\rule{2mm}{0mm}&\rule{2mm}{0mm}&\burst&\burst&\rule{2mm}{0mm}&\rule{2mm}{0mm}&\rule{2mm}{0mm}&\rule{2mm}{0mm}&\rule{2mm}{0mm}&\rule{2mm}{0mm}&\rule{2mm}{0mm}&\rule{2mm}{0mm}&\rule{2mm}{0mm}&\rule{2mm}{0mm}&\rule{2mm}{0mm}&\rule{2mm}{0mm}&\rule{2mm}{0mm}&\rule{2mm}{0mm}&\rule{2mm}{0mm}&\rule{2mm}{0mm}&\rule{2mm}{0mm}&\rule{2mm}{0mm}&\rule{2mm}{0mm}&\rule{2mm}{0mm}\\
 loan~~ & \rule{2mm}{0mm}&\rule{2mm}{0.2mm}&\rule{2mm}{0.4mm}&\rule{2mm}{0mm}&\rule{2mm}{0.333333333333333mm}&\rule{2mm}{0.133333333333333mm}&\rule{2mm}{0.133333333333333mm}&\rule{2mm}{0.266666666666667mm}&\rule{2mm}{0.2mm}&\rule{2mm}{0mm}&\rule{2mm}{0.266666666666667mm}&\rule{2mm}{0.133333333333333mm}&\rule{2mm}{0mm}&\rule{2mm}{0.2mm}&\rule{2mm}{0.333333333333333mm}&\rule{2mm}{0mm}&\rule{2mm}{0.266666666666667mm}&\burst&\rule{2mm}{0.266666666666667mm}&\rule{2mm}{0.133333333333333mm}&\rule{2mm}{0.2mm}&\rule{2mm}{0.2mm}&\rule{2mm}{0mm}&\rule{2mm}{0.2mm}&\rule{2mm}{0.2mm}&\rule{2mm}{0.6mm}&\rule{2mm}{0.2mm}&\rule{2mm}{0.2mm}&\rule{2mm}{0.266666666666667mm}&\rule{2mm}{0.2mm}&\rule{2mm}{0.133333333333333mm}&\rule{2mm}{0mm}&\rule{2mm}{0.2mm}&\rule{2mm}{0.4mm}&\rule{2mm}{0.533333333333333mm}&\rule{2mm}{0.733333333333333mm}&\rule{2mm}{0.333333333333333mm}&\rule{2mm}{0.4mm}&\rule{2mm}{0.6mm}&\rule{2mm}{0.866666666666667mm}&\rule{2mm}{0.533333333333333mm}&\rule{2mm}{0.466666666666667mm}&\rule{2mm}{0.466666666666667mm}&\rule{2mm}{0.333333333333333mm}&\rule{2mm}{0.466666666666667mm}&\rule{2mm}{0.333333333333333mm}&\rule{2mm}{0.4mm}&\rule{2mm}{0.533333333333333mm}&\rule{2mm}{0.4mm}&\rule{2mm}{0.333333333333333mm}&\rule{2mm}{0.2mm}&\rule{2mm}{0mm}\\
 merchant~~ & \rule{2mm}{0mm}&\rule{2mm}{0.133333333333333mm}&\rule{2mm}{0.133333333333333mm}&\rule{2mm}{0mm}&\burst&\rule{2mm}{0.333333333333333mm}&\burst&\burst&\rule{2mm}{0.133333333333333mm}&\rule{2mm}{0.133333333333333mm}&\rule{2mm}{0mm}&\rule{2mm}{0mm}&\rule{2mm}{0.333333333333333mm}&\rule{2mm}{0mm}&\rule{2mm}{0.466666666666667mm}&\rule{2mm}{0.4mm}&\rule{2mm}{0.266666666666667mm}&\rule{2mm}{0.133333333333333mm}&\rule{2mm}{0mm}&\rule{2mm}{0mm}&\rule{2mm}{0mm}&\rule{2mm}{0mm}&\rule{2mm}{0mm}&\rule{2mm}{0mm}&\rule{2mm}{0.133333333333333mm}&\rule{2mm}{0.2mm}&\rule{2mm}{0mm}&\rule{2mm}{0.2mm}&\rule{2mm}{0mm}&\rule{2mm}{0mm}&\rule{2mm}{0mm}&\rule{2mm}{0.333333333333333mm}&\rule{2mm}{0mm}&\rule{2mm}{0mm}&\rule{2mm}{0mm}&\rule{2mm}{0.266666666666667mm}&\rule{2mm}{0mm}&\rule{2mm}{0mm}&\rule{2mm}{0.133333333333333mm}&\rule{2mm}{0mm}&\rule{2mm}{0mm}&\rule{2mm}{0mm}&\rule{2mm}{0mm}&\rule{2mm}{0mm}&\rule{2mm}{0mm}&\rule{2mm}{0.333333333333333mm}&\rule{2mm}{0mm}&\rule{2mm}{0mm}&\rule{2mm}{0mm}&\rule{2mm}{0.466666666666667mm}&\rule{2mm}{0.2mm}&\rule{2mm}{0.2mm}\\
 mortgage~~ & \rule{2mm}{0mm}&\rule{2mm}{0mm}&\rule{2mm}{0.133333333333333mm}&\rule{2mm}{0mm}&\rule{2mm}{0.2mm}&\rule{2mm}{0mm}&\rule{2mm}{0.133333333333333mm}&\rule{2mm}{0.2mm}&\rule{2mm}{0.266666666666667mm}&\rule{2mm}{0mm}&\rule{2mm}{0.133333333333333mm}&\rule{2mm}{0mm}&\rule{2mm}{0mm}&\rule{2mm}{0.2mm}&\rule{2mm}{0.533333333333333mm}&\rule{2mm}{0.333333333333333mm}&\rule{2mm}{0.733333333333333mm}&\rule{2mm}{1.33333333333333mm}&\rule{2mm}{0.533333333333333mm}&\rule{2mm}{0.533333333333333mm}&\rule{2mm}{0.133333333333333mm}&\rule{2mm}{0.133333333333333mm}&\rule{2mm}{0.466666666666667mm}&\rule{2mm}{0.466666666666667mm}&\rule{2mm}{0.133333333333333mm}&\rule{2mm}{0.666666666666667mm}&\rule{2mm}{0.266666666666667mm}&\rule{2mm}{0.333333333333333mm}&\rule{2mm}{0.466666666666667mm}&\rule{2mm}{0.533333333333333mm}&\rule{2mm}{0.2mm}&\rule{2mm}{0.2mm}&\rule{2mm}{0.466666666666667mm}&\rule{2mm}{0.866666666666667mm}&\burst&\rule{2mm}{0.733333333333333mm}&\rule{2mm}{0.8mm}&\rule{2mm}{1.2mm}&\burst&\rule{2mm}{1.53333333333333mm}&\rule{2mm}{1.8mm}&\rule{2mm}{0.733333333333333mm}&\rule{2mm}{0.733333333333333mm}&\rule{2mm}{0.866666666666667mm}&\rule{2mm}{0.6mm}&\rule{2mm}{0.666666666666667mm}&\rule{2mm}{0.8mm}&\rule{2mm}{0.933333333333333mm}&\rule{2mm}{0.466666666666667mm}&\rule{2mm}{0.333333333333333mm}&\rule{2mm}{0.533333333333333mm}&\rule{2mm}{0mm}\\
 nicki~~ & \rule{2mm}{0mm}&\rule{2mm}{0mm}&\rule{2mm}{0mm}&\rule{2mm}{0mm}&\rule{2mm}{0mm}&\rule{2mm}{0mm}&\rule{2mm}{0mm}&\rule{2mm}{0mm}&\rule{2mm}{0mm}&\rule{2mm}{0mm}&\rule{2mm}{0mm}&\rule{2mm}{0mm}&\rule{2mm}{0mm}&\rule{2mm}{0mm}&\rule{2mm}{0mm}&\rule{2mm}{0mm}&\rule{2mm}{0mm}&\rule{2mm}{0mm}&\rule{2mm}{0mm}&\rule{2mm}{0mm}&\rule{2mm}{0mm}&\rule{2mm}{0mm}&\rule{2mm}{0mm}&\rule{2mm}{0mm}&\rule{2mm}{0mm}&\rule{2mm}{0mm}&\rule{2mm}{0mm}&\rule{2mm}{0mm}&\rule{2mm}{0mm}&\rule{2mm}{0mm}&\burst&\burst&\rule{2mm}{0mm}&\rule{2mm}{0mm}&\rule{2mm}{0mm}&\rule{2mm}{0mm}&\rule{2mm}{0mm}&\rule{2mm}{0mm}&\rule{2mm}{0mm}&\rule{2mm}{0mm}&\rule{2mm}{0mm}&\rule{2mm}{0mm}&\rule{2mm}{0mm}&\rule{2mm}{0mm}&\rule{2mm}{0mm}&\rule{2mm}{0mm}&\rule{2mm}{0mm}&\rule{2mm}{0mm}&\rule{2mm}{0mm}&\rule{2mm}{0mm}&\rule{2mm}{0mm}&\rule{2mm}{0mm}\\
 nigeria~~ & \rule{2mm}{0mm}&\rule{2mm}{0mm}&\rule{2mm}{0mm}&\rule{2mm}{0mm}&\rule{2mm}{0mm}&\rule{2mm}{0.2mm}&\rule{2mm}{0mm}&\rule{2mm}{0mm}&\rule{2mm}{0mm}&\rule{2mm}{0mm}&\rule{2mm}{0mm}&\rule{2mm}{0mm}&\rule{2mm}{0mm}&\rule{2mm}{0mm}&\rule{2mm}{0mm}&\rule{2mm}{0mm}&\rule{2mm}{0.333333333333333mm}&\burst&\rule{2mm}{0.466666666666667mm}&\rule{2mm}{0.466666666666667mm}&\rule{2mm}{0.4mm}&\rule{2mm}{0mm}&\rule{2mm}{0.333333333333333mm}&\rule{2mm}{0.533333333333333mm}&\rule{2mm}{0.466666666666667mm}&\rule{2mm}{0.666666666666667mm}&\rule{2mm}{0mm}&\rule{2mm}{0.4mm}&\rule{2mm}{0.666666666666667mm}&\rule{2mm}{0.333333333333333mm}&\rule{2mm}{0.266666666666667mm}&\rule{2mm}{0.8mm}&\rule{2mm}{0.333333333333333mm}&\rule{2mm}{0.266666666666667mm}&\rule{2mm}{0.133333333333333mm}&\rule{2mm}{0.266666666666667mm}&\rule{2mm}{0.466666666666667mm}&\rule{2mm}{0.466666666666667mm}&\rule{2mm}{0.333333333333333mm}&\rule{2mm}{0mm}&\rule{2mm}{0.6mm}&\rule{2mm}{0.266666666666667mm}&\rule{2mm}{0.266666666666667mm}&\rule{2mm}{0.933333333333333mm}&\rule{2mm}{0.333333333333333mm}&\rule{2mm}{0.266666666666667mm}&\rule{2mm}{0.6mm}&\rule{2mm}{0.466666666666667mm}&\rule{2mm}{0.533333333333333mm}&\rule{2mm}{0.6mm}&\rule{2mm}{0.266666666666667mm}&\rule{2mm}{0.133333333333333mm}\\
 pornstars~~ & \rule{2mm}{0mm}&\rule{2mm}{0mm}&\rule{2mm}{0mm}&\rule{2mm}{0mm}&\rule{2mm}{0mm}&\rule{2mm}{0mm}&\rule{2mm}{0mm}&\rule{2mm}{0mm}&\rule{2mm}{0mm}&\rule{2mm}{0mm}&\rule{2mm}{0mm}&\rule{2mm}{0.133333333333333mm}&\rule{2mm}{0mm}&\rule{2mm}{0mm}&\rule{2mm}{0mm}&\rule{2mm}{0.2mm}&\rule{2mm}{0mm}&\rule{2mm}{0mm}&\rule{2mm}{0mm}&\rule{2mm}{0mm}&\rule{2mm}{0mm}&\rule{2mm}{0mm}&\rule{2mm}{0mm}&\rule{2mm}{0mm}&\rule{2mm}{0.133333333333333mm}&\rule{2mm}{0mm}&\rule{2mm}{0mm}&\rule{2mm}{0mm}&\rule{2mm}{0mm}&\rule{2mm}{0mm}&\burst&\rule{2mm}{0mm}&\rule{2mm}{0mm}&\rule{2mm}{0mm}&\rule{2mm}{0mm}&\rule{2mm}{0mm}&\rule{2mm}{0mm}&\rule{2mm}{0mm}&\rule{2mm}{0.133333333333333mm}&\rule{2mm}{0mm}&\rule{2mm}{0mm}&\rule{2mm}{0mm}&\rule{2mm}{0mm}&\rule{2mm}{0mm}&\rule{2mm}{0mm}&\rule{2mm}{0mm}&\rule{2mm}{0mm}&\rule{2mm}{0mm}&\rule{2mm}{0mm}&\rule{2mm}{0mm}&\rule{2mm}{0mm}&\rule{2mm}{0mm}\\
 postmaster~~ & \rule{2mm}{0mm}&\rule{2mm}{0mm}&\rule{2mm}{0mm}&\rule{2mm}{0mm}&\rule{2mm}{0mm}&\rule{2mm}{0mm}&\rule{2mm}{0mm}&\rule{2mm}{0mm}&\rule{2mm}{0mm}&\rule{2mm}{0mm}&\rule{2mm}{0mm}&\rule{2mm}{0mm}&\rule{2mm}{0mm}&\rule{2mm}{0mm}&\rule{2mm}{0mm}&\rule{2mm}{0mm}&\burst&\burst&\rule{2mm}{0.2mm}&\rule{2mm}{0mm}&\rule{2mm}{0mm}&\rule{2mm}{0mm}&\rule{2mm}{0mm}&\rule{2mm}{0mm}&\rule{2mm}{0mm}&\rule{2mm}{0.2mm}&\rule{2mm}{0.333333333333333mm}&\rule{2mm}{0.133333333333333mm}&\rule{2mm}{0mm}&\rule{2mm}{0.333333333333333mm}&\rule{2mm}{0.266666666666667mm}&\rule{2mm}{0.2mm}&\rule{2mm}{0mm}&\rule{2mm}{0mm}&\rule{2mm}{0mm}&\rule{2mm}{0.266666666666667mm}&\rule{2mm}{0.133333333333333mm}&\rule{2mm}{0.133333333333333mm}&\rule{2mm}{0mm}&\rule{2mm}{0mm}&\rule{2mm}{0mm}&\rule{2mm}{0mm}&\rule{2mm}{0mm}&\rule{2mm}{0mm}&\rule{2mm}{0mm}&\rule{2mm}{0mm}&\rule{2mm}{0mm}&\rule{2mm}{0mm}&\rule{2mm}{0mm}&\rule{2mm}{0mm}&\rule{2mm}{0mm}&\rule{2mm}{0mm}\\
 promotion~~ & \rule{2mm}{0mm}&\rule{2mm}{0mm}&\rule{2mm}{0mm}&\rule{2mm}{0.133333333333333mm}&\rule{2mm}{0.133333333333333mm}&\rule{2mm}{0mm}&\rule{2mm}{0mm}&\rule{2mm}{0mm}&\rule{2mm}{0mm}&\rule{2mm}{0mm}&\rule{2mm}{0mm}&\rule{2mm}{0mm}&\rule{2mm}{0.133333333333333mm}&\rule{2mm}{0mm}&\rule{2mm}{0mm}&\rule{2mm}{0mm}&\rule{2mm}{0.133333333333333mm}&\rule{2mm}{0.333333333333333mm}&\rule{2mm}{0.133333333333333mm}&\rule{2mm}{0mm}&\rule{2mm}{0mm}&\rule{2mm}{0mm}&\rule{2mm}{0mm}&\rule{2mm}{0.2mm}&\rule{2mm}{0mm}&\rule{2mm}{0.133333333333333mm}&\rule{2mm}{0mm}&\rule{2mm}{0mm}&\rule{2mm}{0.2mm}&\rule{2mm}{0mm}&\rule{2mm}{0mm}&\rule{2mm}{0mm}&\rule{2mm}{0.333333333333333mm}&\rule{2mm}{0.333333333333333mm}&\rule{2mm}{0.4mm}&\rule{2mm}{0.266666666666667mm}&\rule{2mm}{0.333333333333333mm}&\rule{2mm}{0mm}&\rule{2mm}{0.333333333333333mm}&\rule{2mm}{0.466666666666667mm}&\rule{2mm}{0.333333333333333mm}&\rule{2mm}{0.2mm}&\rule{2mm}{0.2mm}&\rule{2mm}{0.4mm}&\rule{2mm}{0.2mm}&\rule{2mm}{0mm}&\rule{2mm}{0mm}&\rule{2mm}{0mm}&\rule{2mm}{0.133333333333333mm}&\burst&\burst&\rule{2mm}{0.933333333333333mm}\\
 removed~~ & \rule{2mm}{0.6mm}&\rule{2mm}{0.466666666666667mm}&\burst&\rule{2mm}{0.933333333333333mm}&\rule{2mm}{1.26666666666667mm}&\rule{2mm}{1.53333333333333mm}&\rule{2mm}{1.4mm}&\rule{2mm}{1.93333333333333mm}&\rule{2mm}{1.73333333333333mm}&\rule{2mm}{1.6mm}&\rule{2mm}{2mm}&\burst&\burst&\rule{2mm}{2mm}&\rule{2mm}{2mm}&\rule{2mm}{2mm}&\rule{2mm}{2mm}&\rule{2mm}{2mm}&\rule{2mm}{2mm}&\burst&\rule{2mm}{2mm}&\rule{2mm}{2mm}&\rule{2mm}{2mm}&\rule{2mm}{2mm}&\rule{2mm}{2mm}&\rule{2mm}{2mm}&\rule{2mm}{2mm}&\rule{2mm}{2mm}&\rule{2mm}{2mm}&\rule{2mm}{2mm}&\rule{2mm}{2mm}&\rule{2mm}{2mm}&\rule{2mm}{2mm}&\rule{2mm}{2mm}&\rule{2mm}{1.66666666666667mm}&\rule{2mm}{2mm}&\rule{2mm}{2mm}&\rule{2mm}{2mm}&\rule{2mm}{2mm}&\rule{2mm}{2mm}&\rule{2mm}{2mm}&\rule{2mm}{2mm}&\rule{2mm}{2mm}&\rule{2mm}{2mm}&\rule{2mm}{2mm}&\rule{2mm}{2mm}&\rule{2mm}{2mm}&\burst&\burst&\burst&\rule{2mm}{2mm}&\burst\\
 viagra~~ & \rule{2mm}{0mm}&\rule{2mm}{0mm}&\rule{2mm}{0mm}&\rule{2mm}{0mm}&\rule{2mm}{0mm}&\rule{2mm}{0mm}&\rule{2mm}{0.333333333333333mm}&\rule{2mm}{0.333333333333333mm}&\rule{2mm}{0.333333333333333mm}&\rule{2mm}{0.266666666666667mm}&\rule{2mm}{0.2mm}&\burst&\rule{2mm}{0.2mm}&\rule{2mm}{0.266666666666667mm}&\rule{2mm}{0.466666666666667mm}&\rule{2mm}{0mm}&\rule{2mm}{0.133333333333333mm}&\rule{2mm}{0.333333333333333mm}&\rule{2mm}{0.4mm}&\rule{2mm}{0.266666666666667mm}&\rule{2mm}{0mm}&\rule{2mm}{0.333333333333333mm}&\rule{2mm}{0mm}&\rule{2mm}{0.266666666666667mm}&\rule{2mm}{0.2mm}&\rule{2mm}{0mm}&\rule{2mm}{0mm}&\rule{2mm}{0.133333333333333mm}&\rule{2mm}{0.133333333333333mm}&\rule{2mm}{0.133333333333333mm}&\rule{2mm}{0mm}&\rule{2mm}{0mm}&\rule{2mm}{0.2mm}&\rule{2mm}{0mm}&\rule{2mm}{0.466666666666667mm}&\rule{2mm}{0mm}&\rule{2mm}{0mm}&\rule{2mm}{0mm}&\rule{2mm}{0mm}&\rule{2mm}{0.333333333333333mm}&\rule{2mm}{0.2mm}&\rule{2mm}{0.466666666666667mm}&\rule{2mm}{0.4mm}&\rule{2mm}{0.133333333333333mm}&\rule{2mm}{0.333333333333333mm}&\rule{2mm}{0.333333333333333mm}&\rule{2mm}{0.4mm}&\rule{2mm}{0mm}&\rule{2mm}{0.133333333333333mm}&\rule{2mm}{0mm}&\rule{2mm}{0mm}&\rule{2mm}{0mm}\\
 videos~~ & \rule{2mm}{0mm}&\rule{2mm}{0mm}&\rule{2mm}{0mm}&\rule{2mm}{0mm}&\rule{2mm}{0.133333333333333mm}&\rule{2mm}{0.2mm}&\rule{2mm}{0mm}&\rule{2mm}{0.266666666666667mm}&\rule{2mm}{0mm}&\rule{2mm}{0.133333333333333mm}&\rule{2mm}{0.333333333333333mm}&\rule{2mm}{0.533333333333333mm}&\rule{2mm}{0.4mm}&\rule{2mm}{0.666666666666667mm}&\rule{2mm}{0.733333333333333mm}&\rule{2mm}{1mm}&\rule{2mm}{0.6mm}&\rule{2mm}{0.4mm}&\rule{2mm}{0.533333333333333mm}&\rule{2mm}{0.2mm}&\rule{2mm}{0.4mm}&\rule{2mm}{0.466666666666667mm}&\rule{2mm}{0.733333333333333mm}&\rule{2mm}{0.8mm}&\rule{2mm}{0.533333333333333mm}&\rule{2mm}{0.333333333333333mm}&\rule{2mm}{1.06666666666667mm}&\rule{2mm}{0.733333333333333mm}&\rule{2mm}{0.666666666666667mm}&\rule{2mm}{1mm}&\burst&\rule{2mm}{1.33333333333333mm}&\rule{2mm}{0.933333333333333mm}&\rule{2mm}{0.4mm}&\rule{2mm}{0.666666666666667mm}&\rule{2mm}{0.466666666666667mm}&\rule{2mm}{1.53333333333333mm}&\rule{2mm}{1.26666666666667mm}&\rule{2mm}{1.53333333333333mm}&\rule{2mm}{0.8mm}&\rule{2mm}{0.6mm}&\rule{2mm}{1.2mm}&\rule{2mm}{0.8mm}&\rule{2mm}{0.133333333333333mm}&\rule{2mm}{0.2mm}&\rule{2mm}{0.266666666666667mm}&\rule{2mm}{0.266666666666667mm}&\rule{2mm}{0mm}&\rule{2mm}{0.2mm}&\rule{2mm}{0.2mm}&\rule{2mm}{0mm}&\rule{2mm}{0.2mm}\\
 virus~~ & \rule{2mm}{0mm}&\rule{2mm}{0mm}&\rule{2mm}{0mm}&\rule{2mm}{0mm}&\rule{2mm}{0mm}&\rule{2mm}{0mm}&\rule{2mm}{0mm}&\rule{2mm}{0mm}&\rule{2mm}{0mm}&\rule{2mm}{0mm}&\rule{2mm}{0mm}&\rule{2mm}{0mm}&\rule{2mm}{0mm}&\rule{2mm}{0mm}&\rule{2mm}{0mm}&\rule{2mm}{0mm}&\burst&\burst&\rule{2mm}{0.133333333333333mm}&\rule{2mm}{0.133333333333333mm}&\rule{2mm}{0mm}&\rule{2mm}{0mm}&\rule{2mm}{0mm}&\rule{2mm}{0mm}&\rule{2mm}{0mm}&\rule{2mm}{0.4mm}&\rule{2mm}{0.333333333333333mm}&\rule{2mm}{0.133333333333333mm}&\rule{2mm}{0.133333333333333mm}&\rule{2mm}{0.266666666666667mm}&\rule{2mm}{0.133333333333333mm}&\rule{2mm}{0mm}&\rule{2mm}{0.133333333333333mm}&\rule{2mm}{0mm}&\rule{2mm}{0.133333333333333mm}&\rule{2mm}{0mm}&\rule{2mm}{0mm}&\rule{2mm}{0mm}&\rule{2mm}{0mm}&\rule{2mm}{0.133333333333333mm}&\rule{2mm}{0mm}&\rule{2mm}{0mm}&\rule{2mm}{0.2mm}&\rule{2mm}{0.266666666666667mm}&\rule{2mm}{0.266666666666667mm}&\rule{2mm}{0.2mm}&\rule{2mm}{0mm}&\rule{2mm}{0.4mm}&\rule{2mm}{0mm}&\rule{2mm}{0mm}&\rule{2mm}{0mm}&\rule{2mm}{0mm}\\
 weight~~ & \rule{2mm}{0.133333333333333mm}&\rule{2mm}{0mm}&\rule{2mm}{0mm}&\rule{2mm}{0mm}&\rule{2mm}{0.133333333333333mm}&\rule{2mm}{0mm}&\rule{2mm}{0mm}&\rule{2mm}{0.2mm}&\rule{2mm}{0.2mm}&\rule{2mm}{0mm}&\rule{2mm}{0.133333333333333mm}&\rule{2mm}{0mm}&\rule{2mm}{0mm}&\rule{2mm}{0.133333333333333mm}&\rule{2mm}{0mm}&\rule{2mm}{0.2mm}&\rule{2mm}{0mm}&\rule{2mm}{0mm}&\rule{2mm}{0.133333333333333mm}&\rule{2mm}{0.133333333333333mm}&\rule{2mm}{0.2mm}&\rule{2mm}{0.333333333333333mm}&\rule{2mm}{0.133333333333333mm}&\rule{2mm}{0.133333333333333mm}&\rule{2mm}{0.133333333333333mm}&\rule{2mm}{0mm}&\rule{2mm}{0.133333333333333mm}&\rule{2mm}{0.333333333333333mm}&\rule{2mm}{0.2mm}&\rule{2mm}{0.333333333333333mm}&\rule{2mm}{0.2mm}&\rule{2mm}{0.133333333333333mm}&\rule{2mm}{0.533333333333333mm}&\rule{2mm}{0.4mm}&\rule{2mm}{0.266666666666667mm}&\rule{2mm}{0.133333333333333mm}&\rule{2mm}{0.6mm}&\rule{2mm}{0.6mm}&\rule{2mm}{0.2mm}&\burst&\burst&\rule{2mm}{0.8mm}&\rule{2mm}{0.333333333333333mm}&\rule{2mm}{0.4mm}&\rule{2mm}{0.4mm}&\rule{2mm}{0.533333333333333mm}&\rule{2mm}{0.533333333333333mm}&\rule{2mm}{0.6mm}&\rule{2mm}{0.4mm}&\rule{2mm}{0.533333333333333mm}&\rule{2mm}{0.4mm}&\rule{2mm}{0.4mm}\\
\hline
\end{tabular}

    \caption{Frequency and burstiness of spam terms.}
    \label{fig:chi2}
  \end{figure*}
}

Section~\ref{sect:class-skew} made the case that the amount of spam drifts
over time, so class distributions vary.  It is also true that the
\emph{content} of spam changes over time, so class-conditioned feature
probabilities will change as well.

Some spam topics are perpetual, such as advertisements for pornography sites,
offers for mortgage re-financing, and moneymaking schemes.  Other topics are
bursty or occur in epidemics.

One notorious example of a spam ploy coming into vogue is the ``Nigerian
Money'' scam, a get-rich-quick scam in which help was solicited to transfer
money from a Nigerian bank account \cite{Sturgeon:2003b}.  The details varied,
but the sender usually claimed to be responsible for a large bank account and
requested assistance in ``liberating'' the funds from the Nigerian government.
The sender was willing to pay generously for access to a foreign bank account
into which the money would be transfered.  This account was usually drained of
funds once access was granted.  Eventually the people responsible for the scam
were arrested, and spam of that type declined quickly (unfortunately, variants
continue to circulate as other people adopt the general idea).  Prior to this
scam, keywords such as \texttt{nigeria} and \texttt{assistance} were not
strong predictors of spam.

A more dramatic episode occurred in April of 2003 when decks of
playing cards depicting ``Iraq's Most Wanted'' were made available for sale.
These cards were advertised primarily via spam.  The advertising campaign
created such a spam blizzard that its story---and the campaign's success
---were written up in the New York Times \cite{Hansell:2003}.  This campaign
abated quickly and few of the terms uniquely associated with this episode
retain much predictive power now.

The point for researchers is that spam content changes so the ``spam''
concept should drift inevitably.  Some components (disjuncts) of the
concept description should remain constant or change only slowly.
Others will spike during epidemics, as specific scams or merchandising
schemes come into vogue.  Even perpetual topics do not exhibit
constant term frequencies.

It is difficult to estimate how much we can expect ``spam'' as a
concept to drift over time, in part because no metric of concept drift
has been adopted by the community.  It is beyond the scope of this
paper to present a rigorous investigation of concept drift in spam,
but a simple technique can demonstrate significant word frequency
variation.

Swan and Allan \cite{SwanAllan:1999} employed a $\chi^2$ test to
discover ``bursty'' topics in daily news stories.  Their test was
designed to determine whether the appearance of a term on a given day
was statistically significant.  This test can be applied to weekly
groups of spam messages in Wouters' archive, using the average weekly
frequency of a term as its expected value.  The results are shown in
figure~\ref{fig:chi2}, with selected terms listed on the left side and
a column for every week (1--52) of 2002 extending to the right.  The
height of a bar at a term-week is proportional to the term's frequency
in that week.  The special symbol ``\burst'' denotes a term burst: it
appears in a term's row if that term appeared more than four times in
the week and the $\chi^2$ test succeeded at $p<0.01$.

Figure~\ref{fig:chi2} shows that spam has complex time-varying behavior.  Some
terms recur intermittently, such as \textsf{adult, click, free, hot} and
\textsf{removed}.  Others are episodic, \eg\ the terms common in a ``Nigerian
scam'' burst in week 18 (\textsf{nigeria, lagos, assistance, beneficiary}) and
terms in a ``pornstar videos'' burst in weeks 31--32 (\textsf{0rgy, awesome,
  pornstars, jenna, lauren, nicki}).  The term \textsf{christmas} bursts 
late in the year and presumably reappears every year around the same time.




Spam behavior is not simply a matter of one concept drifting to
another in succession, but instead is a superimposition of constant,
periodic and episodic phenomena.  Researchers in data mining have
studied classification under concept drift but it remains an open
problem.  Work in Topic Detection and Tracking \cite{Allan:2002} is
likely to be relevant to spam classification, though technically it
addresses a different problem.  No detailed study of real-world
concept drift has yet been undertaken, and to the best of our
knowledge there are no standard datasets for studying it.  A
longitudinal spam dataset would be an excellent testbed for
investigating issues in concept drift and stream classification.

\subsection{Intelligent adaptive adversaries}
\label{sec:intell-adapt-advers}

The spam stream changes over time as different products or scams, marketed by
spam, come into vogue.  There is a separate reason for concept drift: spammers
are engaged in a perpetual ``arms race'' with email filters
\cite{CranorLaMacchia:1998,Willis:2003}.

Over time spammers have become increasingly sophisticated in their
techniques for evading filtering \cite{Machlis:2003}.  In its early
days, spam would have predictable subject lines like \texttt{MAKE
  MONEY FAST!} and \texttt{Re\-fin\-ance your mort\-gage}.  Messages
would be addressed to \texttt{Un\-dis\-closed\_re\-cip\-i\-ents} or
\texttt{nobody}.  As basic header filtering became common in e-mail
clients, these obvious text markers were simple to filter upon so spam
could be discarded easily.  As message body scanning became common,
fragments such as \texttt{vi\-ag\-ra} and \texttt{click here} could be
checked for as well.

To circumvent simple filtering, spammers began to employ content obscuring
techniques such as inserting spurious punctuation, using bogus HTML tags and
adding HTML comments in the middle of words.  It is now common to see
fragments such as these:
\begin{itemize}
  \pagebreak[0]
  \ttfamily
\item v.ia.g.ra
\item 100\% Mo|ney Back Guaran|tee!
\item \sloppy Our pro<br2sd9/>duct is doctor reco<br2sd9 />mmen<br2sd9/>ded
  and made from 100\% natu<br2sd9/>ral ingre<br2sd9/>dients.
\item \sloppy C<!--7udzl53l5spp6-->lic<!--yajiwn1xnbecx2-->k
  he<!--ehc0aj2pvwu-->re</a>
\item In\`cr\"e\"as\"e t\"est\"ost\"er\"on\"e by 254\%

\end{itemize}
When rendered, these are recognizable to most people but they foil simple word
and phrase filtering.  To counter this, some filters remove embedded
punctuation and bogus HTML tags before scanning, and consider them to be
additional evidence that the message is spam.

Spammers are also aware that filters use bayesian word analysis and
content hashing, so they often pepper their messages with common
English words and nonsense words to foil these techniques
\cite{Machlis:2003}.  Messages are designed so these words are
discarded when the text is rendered, or are rendered unobtrusively.
Graham-Cumming \cite{Graham-Cumming:2003} maintains an extensive
catalog of the techniques used by spammers to confuse filters.

Whatever new filtering capabilities arise, it is just a matter of time
before spammers find ways to evade them.  In machine learning terms,
spammers have a strong interest in making the ``spam'' and
``legitimate'' classes indistinguishable.  Because the discrimination
ability of spam filters improves continually, the resulting concept
drifts.

Because of this text distortion, \iv\ spam filtering diverges
significantly from most text classification and information retrieval
problems, where authors are not deliberately trying to obfuscate
content and defy indexing.  Researchers should expect that they will
have to develop techniques unknown in these related fields.  Much of
the effort of developing spam filters will probably shift from feature
combining (\ie\ experimenting with different induction algorithms) to
feature \emph{generation} (\ie\ devising automatic feature generation
methods that can adapt to new distortion patterns).



Such an arms race is not uncommon.  Co-evolving abilities appear often when
access to a desired resource is simultaneously sought and blocked by
intelligent, adaptive parties.  Fraud analysts observe criminals developing
increasingly sophisticated techniques in response to security improvements
\cite{fraud-chapter}.  With spam, the desired resource is the attention of
email users, and spam may be seen as a way of illicitly gaining access to it.
Another example of an arms race occurs in e-commerce.  As pricing schemes have
become more sophisticated, consumers have become more adept at gaming the
systems.  Sophisticated ``shop bots'' have been developed, and on-line
merchants have had to develop ways to keep pricing information from them.
Both sides continue to improve their techniques.  Finally, the well-publicized
conflict between the music swapping networks and the American music industry
shows characteristics of an arms race, as both sides develop more
sophisticated methods in their battle over access to copyrighted material
\cite{Krebs:2003}.

Such co-adaptation of intelligent agents is foreign to most data mining
researchers: the data are mined and the results are deployed, but the data
environment is not considered to be an active entity that will react in turn.
With the internet, much information is freely and automatically available by
all parties, and interactivity is the rule.  I propose that the future will
bring more scenarios involving feedback and co-adaptation.  Data miners may
have to consider the effects of mining on their task environment, and perhaps
incorporate such concerns into the data mining process.  Possible strategies
include concealing one's deployed techniques from adversaries, incorporating
deception into techniques, or simply speeding up the deployment cycle to adapt
more quickly to adversaries' moves.  Spam filtering could be a useful domain
in which to explore such strategies.

\section{Meeting the challenge}

This paper has made the case that \iv\ spam filtering can be a complex data
mining problem with difficult challenging characteristics:
\begin{itemize}
\item Skewed and changing class distributions
\item Unequal and uncertain error costs
\item Complex text patterns requiring sophisticated parsing
\item A disjunctive target concept comprising superimposed phenomena with
  complex temporal characteristics
\item Intelligent, adaptive adversaries
\end{itemize}
Researchers wishing to explore these issues would do well to study \iv\ spam
filtering.  Controlled laboratory datasets exhibiting these characteristics
are often difficult to acquire and to share.  Spam filtering, on the other
hand, is an excellent domain for investigating these problems.

Researchers wishing to pursue this domain should begin collecting
longitudinal data in a controlled manner.  Spam is notoriously easy to
attract.  Several studies have measured the extent to which various
activities attract spam \cite{Lake:2001,CDT:2003}, and this
information may be useful.  It is easy to create ad hoc email
addresses (for example, through Hotmail or Yahoo) and to advertise
them in a controlled manner to attract spam.  Such addresses are
sometimes called ``spam traps'' and are used by email filtering
companies such as BrightMail to obtain a continuous clean feed of spam
for analysis.

A more difficult problem is that of obtaining shareable corpora of non-spam
email, which often contain personal details that people want to keep private.
Two general approaches have been taken:
\begin{enumerate}
  
\item Researchers who have contributed personal email have sought ways to
  anonymize it.  The contributors of the UCI ``spambase'' dataset
  achieved this by reducing the original messages to word frequencies
  and performing feature selection upon the set.  Unfortunately, this
  makes it difficult for other researchers to experiment with
  alternative feature selection or text processing operations on the
  data.

  Androutsopoulos \etal\ \cite{AndroutsopoulosEtal:2000b} have developed a
  basic encoding technique for sharing data without compromising privacy.
  Their software and several of their datasets are available; see Appendix A.
  
\item Androutsopoulos~\etal\ \cite{AndroutsopoulosEtal:2000a} have suggested
  using messages from websites and public mailing lists as proxies for
  personal email.  Their ``Ling-spam'' corpus uses messages from the moderated
  Linguist list.  Other researchers have suggested that such messages may not
  be representative of the email most people receive.  Whether this renders
  mailing list data ineffective for exploring \iv\ spam filtering remains to
  be studied.
  
\end{enumerate}

However researchers decide to generate such corpora, they should consider
making their data publicly available.

This article has outlined the challenges of \iv\ spam filtering and explained
how pursuing such challenges could help data mining.  It is hoped that this
article stimulates interest in the problem.  The appendix and references
should serve as useful resources for researchers wishing to pursue it.

\section*{Acknowledgements}

The opinions in this paper are those of the author and do not necessarily
reflect the policies or priorities of the Hewlett-Packard Corporation.

I wish to thank Melissa McDowell for providing spam and email feeds.  Thanks
to Rob Holte and Chris Drummond for help on cost curves.  Thanks to Julian
Haight for his continuing work on SpamCop and for allowing use of
figure~\ref{fig:spamcop_month}; thanks to Kristian Eide for providing the data
in figure~\ref{fig:keide}; and thanks to Paul Wouters and the people at
SpamArchive.org for making their data publicly available.

Much open source software was used in preparing this paper.  I wish to thank
the authors and maintainers of XEmacs, \LaTeX, Grace, Perl and its many
user-contributed packages, and the Free Software Foundation's GNU Project.

\appendix
\section{Sources of spam data}

There are several sources of spam data on the internet, though researchers
should be aware of their limitations.

\begin{enumerate}
  
\item Several static databases have been used by the machine learning
  community.  The UCI database
  ``spambase''\footnote{\url{ftp://ftp.ics.uci.edu/pub/machine-learning-databases/spambase}}
  has a featurized version of spam and legitimate email.  Androutsopoulos
  \etal\ \cite{AndroutsopoulosEtal:2000a} have made available several of their
  corpora containing both spam and personal email.  All are available for
  download from \url{http://www.iit.demokritos.gr/skel/i-config/}.  Note that
  messages in these databases are not reliably timestamped so they are not
  useful for measuring time-varying aspects of spam.
  
\item Paul Wouters, of Extended Internet, has an extensive archive of spam
  available at \url{http://spamarchive.xtdnet.nl/}.  His archive covers
  several years.  His messages from 2002 were used to produce
  figures~\ref{fig:weekly_volume}a and \ref{fig:chi2}.
  
\item Richard Jones of Annexia.Org has made a longitudinal spam archive
  available at \url{http://www.annexia.org/spam/index.msp}.  Although his
  messages are carefully timestamped, note his explanation about the large
  drop around mid-2002, attributed to deleting a number of old mail accounts.
  For this reason I avoided making inferences about spam volume from his
  dataset.
  
\item \url{SpamArchive.org} is a ``community resource used for testing,
  developing, and benchmarking anti-spam tools.  The goal of this project is
  to provide a large repository of spam that can be used by researchers and
  tool developers.''  Current SpamArchive has over 200K spam messages and
  receives about 5000 messages per day.

\item Bruce Guenter has a longitudinal database of spam available at
  \url{http://www.em.ca/~bruceg/spam/}.  See the caveat below about measuring
  spam volume.

\end{enumerate}

Note that these datasets are archives of spam saved over time and were not
designed to be controlled research datasets.  It is important to understand
the limitations of measuring spam volume from any of them.  They are kept by
owners of entire sites rather than individual accounts so the spam may be
extracted from several mailboxes.  The mailboxes may include \texttt{admin}
and \texttt{webmaster}, which are believed to receive more spam than average.
Some of these administrators even use ``spam trap'' addresses deliberately to
attract spam.  Finally, note that being active on Usenet or the web can often
get a user added to spamming lists---as can making a spam archive available on
the web.  For all of these reasons, these spam archives may contain more spam
than the average email user typically gets.


\bibliographystyle{abbrv}
\bibliography{MLSpamBibliography,fraud,spam,ROC,crossrefs}

\end{document}